\documentclass[runningheads]{llncs}

\usepackage{eccv}

\usepackage{eccvabbrv}

\usepackage{graphicx}
\usepackage{booktabs}
\usepackage{hyperref}

\usepackage[accsupp]{axessibility}  

\usepackage{bm}
\usepackage{multicol}
\usepackage{multirow}
\usepackage{amsmath}
\usepackage{amsfonts}
\usepackage[export]{adjustbox}
\usepackage{comment}
\newcommand{\argmax}{\operatornamewithlimits{argmax}}

\begin{document}

\title{Gated Temporal Diffusion for Stochastic Long-Term Dense Anticipation} 



\author{Olga Zatsarynna\inst{1}*\and
Emad Bahrami\inst{1}*\and
Yazan Abu Farha\inst{2} \and 
Gianpiero Francesca\inst{3} \and
Juergen Gall\inst{1,4}}

\authorrunning{O. Zatsarynna et al.}

\institute{University of Bonn, Germany \\
\and
Birzeit University, Palestine\\
\and
Toyota Motor Europe, Belgium \\
\and
Lamarr Institute for Machine Learning and Artificial Intelligence, Germany \\ *Equal contribution 
}


\maketitle

\vspace{-0.6cm}
\begin{abstract}
Long-term action anticipation has become an important task for many applications such as
autonomous driving and human-robot interaction. Unlike short-term anticipation,
predicting more actions into the future imposes a real challenge with the increasing
uncertainty in longer horizons. While there has been a significant progress in
predicting more actions into the future, most of the proposed methods address the
task in a deterministic setup and ignore the underlying uncertainty. In this paper,
we propose a novel Gated Temporal Diffusion (GTD) network that models the uncertainty of both
the observation and the future predictions. As generator, we
introduce a Gated Anticipation Network (GTAN) to model both observed and unobserved
frames of a video in a mutual representation. On the one hand, using a mutual
representation for past and future allows us to jointly model ambiguities in the
observation and future, while on the other hand GTAN can by design
treat the observed and unobserved parts differently and steer the information flow
between them. Our model achieves state-of-the-art results on the
Breakfast, Assembly101 and 50Salads datasets in both stochastic and deterministic
settings.
\end{abstract}

\footnotetext[2]{Corresponding author: zatsarynna@iai.uni-bonn.de}
\footnotetext[3]{Code: \url{https://github.com/olga-zats/GTDA}{}}

\vspace{-0.8cm}
\section{Introduction}
\begin{figure}[t!]
    \centering
    \includegraphics[width=1.0\columnwidth]{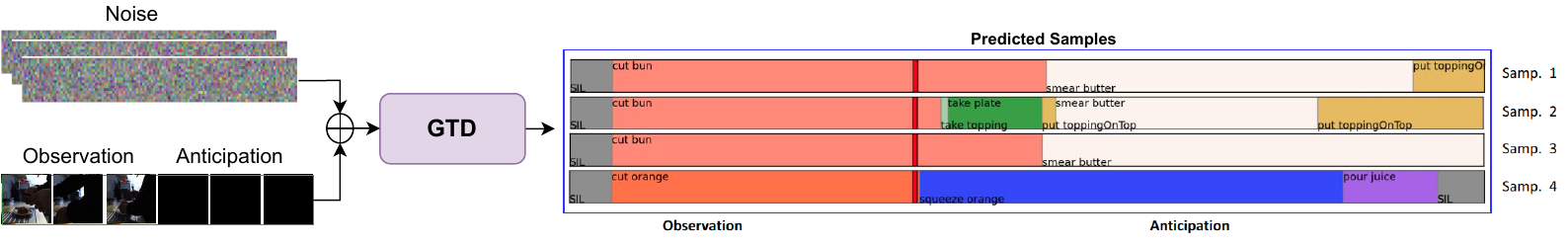}
    \caption{
    The proposed Gated Temporal Diffusion (GTD) model generates multiple future long-term predictions of actions from a single partially observed video. In contrast to previous works, it models the uncertainty of both the observation and the future. In this example, the light conditions make it difficult to distinguish if a bun or an orange is cut. This ambiguity is reflected in the predicted samples where the uncertainty of the past impacts the predicted future.}
    \label{fig:teaser}
    \vspace{-0.7cm}
\end{figure}

\vspace{-0.3cm}
\label{sec:intro}
In this work, we address the task of long-term dense action anticipation. Given a video as observation, the goal is to predict future actions and their durations where the forecasting horizons can span from several seconds to several minutes into the future, which makes it a challenging problem. Yet, solving it is crucial for many real-world applications, such as autonomous driving or human-robot interaction. 
Over the last few years, the task has gained increased attention and there has been a steady progress~\cite{Farha_2018_CVPR, gong2022future, Ke_2019_CVPR, farha2020gcpr, sener2020temporal, nawhal2022anticipatr}, but most works address this task deterministically, which means that only one prediction is made for a single observation.     
The task of forecasting future actions, however, is highly uncertain by nature, especially for longer anticipation horizons, since the same observation can have multiple plausible continuations. Despite its importance, dealing with uncertainty for long-term dense action anticipation has so far received little attention. Farha \textit{et al.}~\cite{farha2019uaaa} proposed to address this task in a stochastic manner. The approach generates multiple predictions in an autoregressive way by predicting the probabilities of the next action and its duration, and then sampling from the predicted probabilities. Alternatively, a GAN-based probabilistic encoder-decoder network has been proposed to generate multiple predictions~\cite{zhao2020async}. Both approaches, however, assume that the action labels of the observed frames are already given, either pre-estimated~\cite{farha2019uaaa} or taken from the ground-truth actions~\cite{farha2019uaaa, zhao2020async}. In this way, the uncertainty of the observation is not taken into account. The observation, however, can also be ambiguous due to occlusions or difficult light conditions as shown in Fig.~\ref{fig:teaser}. We, therefore, argue that stochastic action anticipation needs to consider not only the ambiguity of the future but also of the observation.  

In this work, we propose a diffusion-based model~\cite{sohl2015NonEquTherm, song2019generative, ho2020denoisingDiff, song2021scorebased,ramesh2022hierarchical,chan2024hyperdiffusion} - \textbf{G}ated \textbf{T}emporal \textbf{D}iffusion (GTD) -  that models the uncertainty of both the observation and the future predictions. In particular, we make use of diffusion since it inherently models uncertainty by generating multiple predictions. The key aspect of our GTD, however, is that we explicitly model the uncertainty of observed and future actions jointly by predicting them simultaneously with a single diffusion process. To this end, we use a joint sequence representation illustrated in Fig.~\ref{fig:teaser}, which we construct by extending the observed video frames with zero padding in place of future frames. We then cast this sequence as the per-frame conditioning of the past and future action variables, whose conditional joint probabilities will be learned during training. 
While grouping both observed and unobserved action variables allows us to model them using a single diffusion process, these two parts are nevertheless intrinsically distinct. To take this difference into account, we propose a novel generator - \textbf{G}a\textbf{T}ed \textbf{A}nticipation \textbf{N}etwork (GTAN) - that can differentiate between past and future frames in a distinct manner and steer the information flow between them in a data-driven way.

We evaluate our proposed model on the  
Breakfast~\cite{Kuehne12}, Assembly101~\cite{sener2022assembly101} and 50Salads~\cite{Stein2013CombiningEA} datasets. We show that our approach achieves state-of-the-art results for stochastic long-term dense action anticipation. Additionally, despite our focus on stochastic anticipation, we also demonstrate that our proposed GTAN network achieves state-of-the-art results in the deterministic setting. 
In summary, we make the following contributions:
\vspace{-0.2cm}
\begin{itemize}
    \item We propose the first diffusion model - Gated Temporal Diffusion - for stochastic long-term action anticipation that jointly models observed and future actions.
    \item We propose a novel generator backbone - Gated Anticipation Network - which steers the information flow between observed and future frames. 
    \item We show that our model achieves state-of-the-art results in stochastic and deterministic settings on three datasets: Breakfast, Assembly101 and 50Salads. 
\end{itemize}

\vspace{-0.3cm}
\section{Related Work}
\label{sec:related_work}
\vspace{-0.3cm}

\textbf{Action Anticipation in Videos.}
The task of action anticipation in videos is to forecast future actions given video observations from the current and past moments of time. Following the introduction of anticipation benchmarks on the recent~\cite{Damen2021PAMI, Grauman2021Ego4DAT, sener2022assembly101} and less recent~\cite{Kuehne12, Li2020InTE} datasets, this task has been gaining increasing attention~\cite{zhong2023survey}. In general, existing literature distinguishes between short-term and long-term anticipation approaches, based on the length of the anticipation horizon. Short-term anticipation methods~\cite{Miech_2019_CVPR_Workshops, girdhar2021anticipative, zatsarynna_2022_gcpr, zatsarynna2021MMTCN, zatsarynna2023goal, furnari2020rulstm, Zhong2022AnticipativeFF, Liu_2020_ECCV, zhao2022testra, Wu2020LearningTA, sener2020temporal} forecast only a single action that takes place in the near future. Long-term action anticipation methods, which are the focus of our work, have a longer forecasting horizon and predict multiple future actions several minutes into the future. 
Among long-term anticipation approaches, one can further pin down several established task formulations. 
In the first line of work~\cite{Farha_2018_CVPR}, Farha~\etal proposed to anticipate action classes densely for a subset of frames from the predefined future time interval. This formulation, addressed in~\cite{Farha_2018_CVPR, gong2022future, Ke_2019_CVPR, farha2020gcpr, sener2020temporal, nawhal2022anticipatr} and referred to here as \textit{deterministic long-term dense anticipation}, implies the necessity to predict not only classes of future actions but also their duration. In their work, Farha~\etal~\cite{Farha_2018_CVPR} introduced two models with different modes of predictions based on CNN and RNN networks. Ke~\etal~\cite{Ke_2019_CVPR} proposed a TCN-based model with time conditioning to enable direct action prediction at a predefined moment of time. In the later works, several encoder-decoder architectures were proposed by Farha~\etal~\cite{farha2020gcpr}, Sener~\etal~\cite{sener2020temporal}, Gong~\etal~\cite{gong2022future} and Nawhal~\etal~\cite{nawhal2022anticipatr} based on GRUs~\cite{bahdanau2014neural}, Non-Local blocks~\cite{Wang2017NonlocalNN} and Transformer layers~\cite{vaswani2017attention}. 
In the second line of work~\cite{farha2019uaaa}, Farha~\etal extended the previous formulation into the probabilistic domain, so that the uncertainty of future anticipation could be taken into account. In this formulation, called here \textit{stochastic long-term dense anticipation}, actions still need to be anticipated densely, however, several future sequences are allowed to be predicted for a single observation. In their work~\cite{farha2019uaaa}, the authors proposed a probabilistic RNN network that made autoregressive predictions based on the samples drawn from the predicted action class distributions. Zhao~\etal~\cite{zhao2020async} later proposed a GAN-based probabilistic encoder-decoder network. 
Lastly, in the anticipation frameworks from~\cite{ego-topo, Grauman2021Ego4DAT}, which we refer to as \textit{transcript long-term anticipation}, the estimation of duration for future actions is discarded. This setting has been addressed in~\cite{nawhal2022anticipatr, ego-topo, Grauman2021Ego4DAT, Mascaro_2023_WACV, Ashutosh_2023_CVPR, Das2022VideoC}. 

\textbf{Diffusion}.
Diffusion models~\cite{sohl2015NonEquTherm, song2019generative, ho2020denoisingDiff, song2021scorebased, ramesh2022hierarchical} are a class of deep probabilistic generative models that recover the data sample from Gaussian noise via a gradual denoising process. In a following work, denoising diffusion implicit models (DDIMs)~\cite{song2021denoising} were introduced to speed up the diffusion model sampling.
Later, \cite{chen2023analog, campbell2022discreteDenoising} extended the continuous diffusion models to generate discrete data.
Diffusion models have shown outstanding results in various generation tasks such as image synthesis~\cite{Dhariwal2021beatGan, Rombach2022stableDiff}, video generation~\cite{blattmann2023videoldm, ho2022video}, speech processing~\cite{Popov2021GradTTSAD, Tae2022editTTS, Yang2023soundDiff}, natural language processing~\cite{austin2021structured, Hoogeboom2021argmaxFlow} and motion generation~\cite{Tanke2023socialDiff, zhang2022motiondiffuse}. Furthermore, diffusion models have achieved great success in computer vision perception tasks such as  segmentation~\cite{baranchuk2022labelefficient, brempong2022denoising, chen2023generalist}, object detection~\cite{Chen2023detDiff}, temporal action segmentation~\cite{Liu2023segmDiff} and detection~\cite{Nag2023tadDiff}. The stochastic nature of diffusion models has been leveraged for motion anticipation~\cite{Tanke2023socialDiff, xu2023stochastic, barquero2023belfusion} and procedure planning~\cite{wang2023pdppprojected}.  

\vspace{-0.3cm}
\section{Stochastic Long-Term Dense Anticipation}
\label{sec:task}
\vspace{-0.5cm}
Previously proposed stochastic long-term anticipation approaches~\cite{farha2019uaaa,zhao2020async} assume that the observed video segments share the same format as the future predictions, namely action labels. This assumption, however, overlooks the ambiguity inherent in certain video observations, such as insufficient context due to limited observation duration, challenging lighting conditions, occlusions, and other factors (see Fig.~\ref{fig:teaser} and Fig.~\ref{fig:qualitative_res_pron}). In such cases, the estimated or pre-defined action labels fail to convey the uncertainty associated with observed frames. 

To overcome this limitation, we aim to jointly model the uncertainty of both future and observed actions. To this end, we propose an approach based on diffusion, known for its suitability for modelling uncertainty~\cite{chan2024hyperdiffusion}. Standard diffusion models, however, cannot be directly applied to the task of long-term action anticipation. Thus, we introduce a novel model for stochastic long-term anticipation, termed Gated Temporal Diffusion for Anticipation (GTD), which we describe in detail in Sec.~\ref{sec:probabilistic}. Although we focus on stochastic anticipation, we also show how our method can be used in the deterministic case in Sec.~\ref{sec:deterministic}. Before discussing our proposed approach, we formally define the task of stochastic long-term anticipation.  

Following the popular protocol introduced by \cite{Farha_2018_CVPR}, the observed part and the duration of the future prediction are defined by percentages $\alpha$ and $\beta$ of the entire video.
More precisely, given $\alpha|V|$ observed frames of a video with $|V|$ frames, the goal is to predict action labels for the future $\beta |V|$ frames.
Accordingly, $N_o = \alpha|V|$ is the number of observed frames, $N_a = \beta|V|$ is the number of frames whose action classes we want to anticipate, and $N = N_o + N_a$ is the total number of frames that are considered. Since $\beta$ can go up to $0.5$, the prediction duration can be very long.

While deterministic approaches make only a single most likely future prediction, stochastic approaches consider the uncertainty of the future modelling and generate multiple $(M>1)$ future samples. 

Formally, the task of stochastic long-term anticipation can be formulated as learning to draw samples from the underlying joint probability of per-frame future actions conditioned on the observed video frames:
\begin{align} 
& \hat{Y}^{N_o+1:N} \sim p_\theta(Y^{N_o+1}, \dots, Y^{N} | F)  \\ \label{eq:obs}
& F = (\phi(x^{1}), \dots, \phi(x^{N_o})) \in \mathbb{R}^{N_o \times D},
\end{align}
where $x^k$ is the $k^{th}$ input frame, $\phi$ is the feature extractor network, $Y^{i}$ is the action variable corresponding to $i^{th}$ frame and $Y^{i:j}$ denotes a sequence of variables for frame $i$ to $j$. 

Since in our work, we aim to additionally model the uncertainty present in the video observations, we instead learn to sample $\hat{Y}=\hat{Y}^{1:N}$ from the conditional joint distribution of both future and observed actions.

\vspace{-0.2cm}
\section{Gated Temporal Diffusion for Anticipation}
\label{sec:probabilistic}
\vspace{-0.3cm}
To model uncertainty and perform stochastic action anticipation, we formulate our network as a diffusion model as illustrated in Fig.~\ref{fig:teaser}. While the standard diffusion model, which we describe in Sec.~\ref{sec:diffusion}, serves as a foundation, it cannot be directly applied to the task of long-term action anticipation. Hence, we introduce a novel model, called Gated Temporal Diffusion (GTD), which jointly models the uncertainty of both observed and unobserved events while preserving the inherent difference between the two. We discuss our proposed approach in Sec.~\ref{sec:diff_anticipation}.

\vspace{-0.2cm}
\subsection{Diffusion Model}
\label{sec:diffusion}
\vspace{-0.2cm}
Diffusion models learn to map noise samples $Y_T\sim\mathcal{N}(\mathbf{0}, \mathbf{I})$ to the samples from the data distribution $Y_0\sim q(Y)$ in an iterative manner using a reverse Markov chain process $p_\theta(Y_{0:T})$ with learnable transition parameters $\theta$:
\vspace{-0.2cm}
\begin{align}
    \text{\small$p_\theta(Y_{0:T})$} &= \text{\small $p_\theta(Y_T)\prod_{t=1}^{T}p_\theta(Y_{t-1} | Y_t)$}, \quad \text{\footnotesize $p_\theta(Y_{t-1} | Y_t) = \mathcal{N}(Y_{t-1}; {\mu}_\theta(Y_t, t), \mathbf{\Sigma}(t))$}. 
\end{align}
To learn these parameters, a forward Markov process is defined. It specifies the transitions in the inverse direction by adding Gaussian noise to the data according to a fixed variance schedule $\beta_1, \dots, \beta_T$:
\vspace{-0.2cm}
\begin{align}
    \text{\small$q{(Y_{1:T} | Y_0}) = \prod_{t=1}^{T} q(Y_t | Y_{t-1})$}, \quad
    \text{\small$q(Y_{t} | Y_{t-1}) = \mathcal{N}(Y_{t}; \sqrt{1-\beta_t}Y_{t-1}, \beta_t\mathbf{I})$}.
\end{align}

\paragraph{Training.} Optimization of diffusion models is performed using the variational lower bound on the negative log-likelihood of the data samples where some properties of the forward process are harnessed. Given the Gaussian nature of the forward transition probabilities, one can use the reparametrization trick~\cite{kingma2013auto} to draw samples directly from $Y_0$ by corrupting it following the schedule $\gamma_t$:
    \begin{align}
        \label{eq:x_t}
        Y_t = \sqrt{\gamma_t}Y_0 + \sqrt{(1 - \gamma_t)}{\epsilon}_t,
    \end{align}
    where ${\epsilon_t}\sim\mathcal{N}(\mathbf{0}, \mathbf{I})$, $\gamma_{t}=\prod_{k=1}^{t}(1-\beta_k)$. 
 For training, $Y_t$ is obtained using the forward process~\eqref{eq:x_t} at step $t\sim\mathcal{U}(1, T)$ and a denoising generator network $G_\theta(Y_t, t)= \hat{Y}_{0,t}$ is trained to reverse the noise and predict the reconstruction $\hat{Y}_{0,t}$ of $Y_0$ by minimizing the $l_2$ reconstruction error between them:
\vspace{-0.2cm}
\begin{align}
    L_{diff} = \mathbb{E}_{t\sim\mathcal{U}(1, T), {\epsilon_t}\sim\mathcal{N}(\mathbf{0}, \mathbf{I})} \lVert G_\theta(Y_t, t) - Y_0 \rVert^2.
\end{align}
The key contribution of a diffusion model is the design of the generator network $G_\theta(Y_t, t)$.    
    
\paragraph{Inference.} 
Once trained, sampling from the diffusion model requires following a sequence of denoising steps. In the DDPM~\cite{ho2020denoisingDiff} sampling procedure, inference follows $T$ denoising steps. Starting at step $t=T$, a random sample is drawn from $Y_T\sim \mathcal{N}(\mathbf{0}, \mathbf{I})$ and fed to the generator $G_\theta$ that predicts $\hat{Y}_{0, T}$. 
Assuming that we have already sampled $Y_t$ and reconstructed $\hat{Y}_{0, t}$ at step $t$, we can generate the next sample $Y_{t-1}$ by first approximating ${\epsilon}_t$ by:
\vspace{-0.2cm}
\begin{align}
        \label{eq:eps_t}
        {\hat{\epsilon}}_t = \frac{1}{\sqrt{1 - \gamma_t}}(Y_t - \sqrt{\gamma_t} \hat{Y}_{0, t}).
\end{align}
Since the reverse transition probabilities  $q(Y_{t-1} | Y_t, Y_0)$ become tractable when conditioned on $Y_0$ and can be expressed as Gaussians $\mathcal{N}(Y_{t-1}; \tilde{\mu}_t, \tilde{\beta_t}\mathbf{I})$, we can estimate the parameters of this Gaussian by:
\begin{align}
      \label{eq:mu_t}
      \hat{\mu}_t  = \sqrt{\gamma_{t-1}}\hat{Y}_{0,t}  + \sqrt{1 - \gamma_{t-1} - \tilde{\beta}_t}{\hat{\epsilon}}_t, \quad
      \tilde{\beta}_t = \frac{1 - \gamma_{t-1}}{1 - \gamma_{t}} \beta_{t}
    \end{align}
    and sample $Y_{t-1} \sim \mathcal{N}(Y_{t-1}; \hat{\mu}_t, \tilde{\beta}_t\mathbf{I})$.
    The steps continue until $t=1$, after which $\hat{Y}_{0,1}$ is taken as the final generated sample.
    For an alternative DDIM~\cite{song2021denoising} sampling, variances of the transition probabilities are set to zero during inference, \ie, $\tilde{\beta}_t=0$, making the denoising process deterministic for a particular noise sample $Y_T$. This way, $Y_{t-1}$ is equal to the mean value $Y_{t-1}=\hat{\mu}_t$. The DDIM sampling scheme performs better than DDPM if less denoising steps are used during inference compared to training~\cite{song2021denoising}.  

    \begin{figure*}[t!]
    \centering
    \includegraphics[width=0.7\linewidth]{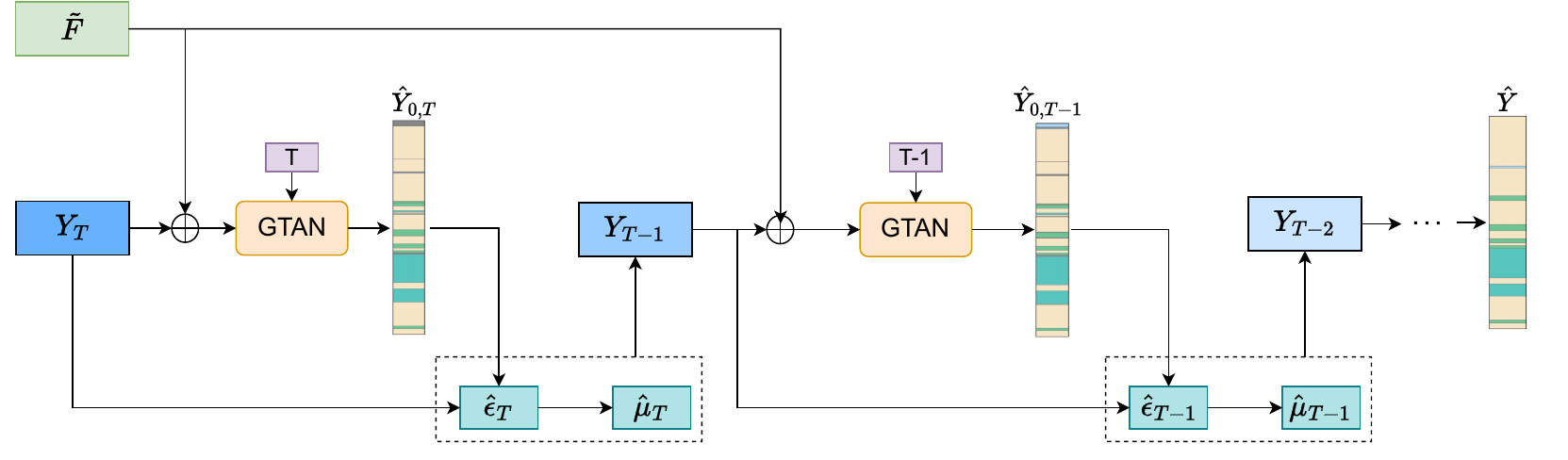}
    \vspace{-0.2cm}
    \caption{We formulate stochastic action anticipation as a diffusion process where the initial input consists of Gaussian noise, $ Y_T $, and zero-padded features, $ \tilde{F} $. Given the inputs, the GTAN generator predicts the denoised action labels, $\hat{Y}_{0, T}$. From step T-1 to 0, the GTAN generator uses self-conditioning by taking the previous denoised action labels as additional input. The noise $\hat{\epsilon}_t$ and mean $\hat{\mu}_{t}$ terms for steps T-1 to 0 are computed using equations~\eqref{eq:eps_t} and~\eqref{eq:mu_t}. $\oplus$ indicates channel-wise concatenation.}
    \label{fig:types_of_training}
    \vspace{-0.35cm}
\end{figure*}
    
\vspace{-0.2cm}
\subsection{Gated Temporal Diffusion for Anticipation}
\label{sec:diff_anticipation}    
\vspace{-0.2cm}

In contrast to previous works, we aim to model uncertainty both in the past observation and the future. We thus extend the formulation described in Sec.~\ref{sec:task} and generate multiple samples not only for the future, but also multiple interpretations of the past as shown in Fig.~\ref{fig:qualitative_res_pron}. Since the uncertainty of the future depends on the uncertainty in the observation, we treat them as a unified sequence $\hat{Y}=\hat{Y}^{1:N}$ and model them with a shared diffusion model. 

While the diffusion model described in Sec.~\ref{sec:diffusion}, generates multiple data samples by sampling $Y_T\sim\mathcal{N}(\mathbf{0}, \mathbf{I})$ repeatedly, it is not directly applicable to our problem for several reasons. Specifically, it operates in the continuous domain and it does not incorporate any conditioning information during the generation process. Moreover, the generator network $G_\theta$ treats all variables uniformly, disregarding the distinction between observed and future action variables in the input sequence. To address these limitations, we propose a Gated Temporal Diffusion model (GTD). As our key contribution, we introduce a GTAN generator network, described in Sec.~\ref{sec:gen}. This network employs gated temporal convolutions and, on one hand, models observed and unobserved data jointly, while, on the other hand, steers the information flow between past and future entries with the learnable gates, thereby controlling their fusion. Apart from that, we also propose discretization and conditioning schemes, elaborated below.             

To model discrete action categories using continuous state diffusion, we represent action labels as one-hot encoded vectors $Y_0 \in \mathbb{R}^{N \times C}$ and regard them as ``analog bits''~\cite{chen2023analog}  that can be directly modelled by continuous state diffusion models. In this way, training remains unchanged and the inference process also remains the same, except that we map generated samples back to the discrete domain by applying the $\argmax$ operation over the class dimension.

To condition the generation on the observed frames, we make use of frame-wise feature vectors $F$ (\ref{eq:obs}) and adapt them to act as the per-frame condition for both observed and future action variables. More specifically, we expand it by incorporating zero-padding to compensate for the absent features of the future unobserved frames:
\vspace{-0.1cm}
\begin{align}\label{eq:F}
    \hat{Y} \sim p_\theta(Y| \tilde{F}), \quad \tilde{F} = (\phi(x^{1}), \dots, \phi(x^{N_o}), \textbf{0}, \dots, \mathbf{0}) \in \mathbb{R}^{N \times D}. 
\end{align}
In this way, the resulting vector $\tilde{F}$ can be used to condition the per-frame diffusion generation process. To this end, we channel-wise concatenate the padded observed frame features $\tilde{F} \in \mathbb{R}^{N \times D} $ to the current sample $Y_t \in \mathbb{R}^{N \times C}$ at each step $t$. Furthermore, we employ self-conditioning~\cite{chen2023analog} by using the previous estimate $\hat{Y}_{0,t+1} \in \mathbb{R}^{N \times C}$ as additional input for the generator $G_\theta$:
     \begin{align}
         G_\theta(Y_t, \hat{Y}_{0,t+1}, \tilde{F}, t) = \hat{Y}_{0,t} \in \mathbb{R}^{N\times C}, \hspace{0.2cm} 
         \vspace{-0.2cm}
     \end{align}
where during training $\hat{Y}_{0,t+1}$ is randomly set to zero with probability $p$, which is then equivalent to training $G_\theta$ without self-conditioning. While $Y_t$, $\hat{Y}_{0,t+1}$, and $\tilde{F}$ are concatenated and used as input to the generator $G_\theta$, shown in Fig.~\ref{fig:types_of_training},
at each diffusion step $t$, we encode the step $t$ as sinusoidal positional embedding. In the next section, we describe our proposed GTAN generator network in detail.    

\vspace{-0.2cm}
\subsection{Gated Anticipation Generator}\label{sec:gen}
\vspace{-0.2cm}
To model temporal dependencies in the input, we use temporal convolutional layers for the generator network $G_\theta$. As the input sequence to our network consists of two distinct parts (observed and unobserved), directly applying the network with classical temporal convolution layers as in other tasks~\cite{AbuFarha2019MSTCNMT, ltc2023bahrami, chinayi2021ASformer} is sub-optimal. This is because the distinction between observed and future parts would not be possible, leading to all values being treated equally. 
While gated convolutions~\cite{Dauphin2016LanguageMW, van2016conditional, aslan2022gtcn, Yu2018FreeFormII} have been proposed for different purposes, we adopt this concept such that the generator network adaptively decides how much mixing between the past and future occurs at different levels of the network.  

\begin{figure*}[t!]
    \centering
    \includegraphics[trim=4mm 0 3mm 0, clip, width=0.9\linewidth]{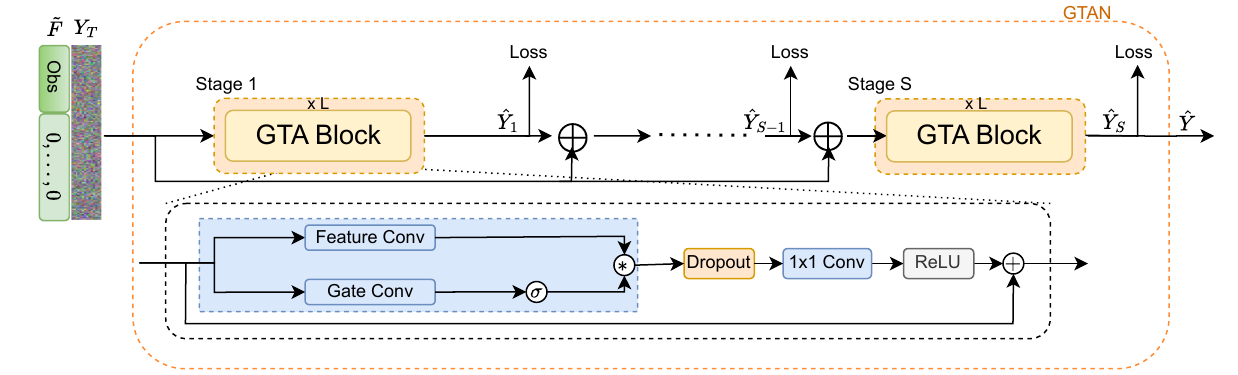}
    \vspace{-0.2cm}
    \caption{The GTAN takes as input a joint representation sequence for observed and future frames. Each stage consists of GTA blocks. The dilated gated convolutions de-activate features at certain frames.}
    \label{fig:arch}
    \vspace{-0.5cm}
\end{figure*}

Motivated by this, we propose a Gated Temporal Anticipation Network (GTAN). It comprises $S$ stages that produce output vectors of action probabilities over $C$ classes. Each stage consists of $L$ residual GTA blocks as shown in Fig.~\ref{fig:arch}. Each GTA block includes two dilated temporal convolutional layers: one for feature processing \textit{(feature convolution)} and another for gating the features \textit{(gate convolution)}. 
Formally, given a feature vector as input, both layers are applied to it separately with the same dilation rate, kernel size, and number of channels. The layers are then combined by the element-wise product between the output of the feature convolution and the sigmoid-activated output of the gate convolution. Given the input feature $\tilde{H}_{s,l-1}$, the output $\bar{H}_{s,l}$ of the gated convolution at layer $l$ and stage $s$ is computed as follows:
\begin{align}
    \bar{H}_{s,l} &= \sigma(W_g * \tilde{H}_{s,l-1}+ b_g) \odot (W_f * \tilde{H}_{s,l-1} + b_f),
\end{align}
where $\sigma$ is the sigmoid function, $*$ denotes the convolution operator, and $\odot$ denotes the element-wise product. $W_g$ and $W_f$ are weights of the convolutional filters and $b_g$ and $b_f$ are biases for gate and feature convolutions, respectively.
The output $\bar{H}_{s,l}$ is then passed through a dropout layer and then to a $1\times1$ convolutional layer followed by the ReLU non-linearity. 
Finally, the output of the layer $l$ is computed by applying a residual connection after the ReLU. At each layer $l$, the dilation factor for the gate and feature convolution is set to $2^l$, \ie, it increases by a factor of two with each layer.

In our ablations studies, we show the benefit of gating and demonstrate the importance of learning temporal gates in a data-driven way, as opposed to using manual masks~\cite{Liu2018ImageIF} or using channel-wise gating~\cite{Hu2017SqueezeandExcitationN}, \ie without the temporal component. We also show that our proposed gated generator is superior to the previously proposed gated temporal convolutional network~\cite{aslan2022gtcn}.

\vspace{-0.2cm}
\subsection{Training and Inference}
\vspace{-0.2cm}
For training our proposed Gated Temporal Diffusion model, we sample step $t\sim\mathcal{U}(1, T)$ and $Y_t$~\eqref{eq:x_t} for a given observation $\tilde{F}$~\eqref{eq:F} with ground-truth $Y_0$. We then apply our proposed GTAN generator $G_\theta(Y_t, \hat{Y}_{S, 0, t+1}, \tilde{F}, t)$ that produces a set of stage-wise predictions $\{\hat{Y}_{s,0,t}\}_{s=1}^{S}$. The self-conditioning with prediction $\hat{Y}_{S, 0, t+1}$ at step $t+1$ is only included if $t < T$. 
We train our network with the $l_2$ reconstruction loss accumulated over all $S$ stages:
\vspace{-0.2cm}
\begin{align}
    L_{stoch} = \mathbb{E}_{t\sim\mathcal{U}(1, T), {\epsilon_t}\sim\mathcal{N}(\mathbf{0}, \mathbf{I})} 
    \sum_{s=1}^{S}\lVert \hat{Y}_{s,0,t} - Y_0 \rVert^2. 
\end{align} 
During inference, we apply the DDIM~\cite{song2021denoising} sampling scheme and use a subset of $D$ denoising steps to get our final predictions. 
As the final output of our model, we take the reconstruction made at the denoising step $t=1$ of the last GTAN stage $S$, \ie, $\hat{Y}=\hat{Y}_{S, 0, 1}$. 
To generate multiple action sequences for a given observation, we run the denoising process starting from $M$ different noise samples $\{Y_{T, m}\}_{m=1}^{M}$, which are then reconstructed into distinct predictions $\{\hat{Y}_{m}\}_{m=1}^{M}$, where $\{\hat{Y}_{m}^{N_o+1:N}\}_{m=1}^{M}$ are the future action predictions.  

\vspace{-0.3cm}
\subsubsection{Deterministic Anticipation.}
\label{sec:training}\label{sec:deterministic} 
\vspace{-0.1cm}
Although we focus on stochastic long-term anticipation, we also report results for the deterministic anticipation. In this case, we use the proposed GTAN without the diffusion process. Given $\tilde{F}$ as the input sequence, our model produces intermediate $\{\hat{Y}_s\}_{s=1}^{S-1}$ and final $\hat{Y}=\hat{Y}_S$ predictions. Using these predictions, we train our network using the standard cross-entropy loss applied to all stages and frames:
\vspace{-0.2cm}
\begin{align}
        L_{determ}= -\sum_{s=1}^{S}\sum_{n=1}^{N}Y^{n} \log\hat{Y}^{n}_{s},
\end{align}
where $Y^{n} \in \mathbb{R}^{C}$ is the one-hot encoded ground-truth action at frame $n$. During inference, we only consider predictions for the future frames.
     
\vspace{-0.2cm}
\section{Experiments}
\vspace{-0.2cm}
\subsection{Datasets and Evaluation Metrics}
\vspace{-0.2cm}
We evaluate our proposed network on three challenging datasets: Breakfast, Assembly101 and 50Salads. 

\textbf{Breakfast}~\cite{Kuehne12} contains breakfast preparation videos. It contains 1712 videos of actors preparing 10 breakfast-related dishes. Each video is densely annotated with actions from 48 classes. On average, videos are $2.3$ minutes long and contain $6$ action segments. The longest video is $10.8$ minutes long, so the duration of anticipation is up to 5.4 minutes. For evaluation, we use the standard 4 splits for cross-validation and report the average result.

\textbf{Assembly101}~\cite{sener2022assembly101} is a large-scale dataset of toy vehicle assembly and disassembly. It contains 4321 videos, which are densely annotated with 100K coarse action segments from 202 coarse action classes. 
The duration of anticipation is up to 12.5 minutes in videos lasting 25 minutes.
The dataset is divided into train, validation, and test splits. Since the test split is not publicly available, we train our model on the train split and report our results on the validation set.

\textbf{50Salads}~\cite{Stein2013CombiningEA} contains videos of salad preparations. It comprises 50 videos annotated with dense segments from 17 fine-grained actions. The mean length of videos is $6.4$ minutes, while the longest video is $10.1$ minutes long, so the duration of anticipation is up to $5.1$ minutes.  For evaluation, we use the predefined 5 splits for cross-validation and report the average result.

For a fair comparison with existing methods on these datasets, we used previously extracted visual features. On the Breakfast and 50Salads datasets, we use the I3D~\cite{Carreira2017QuoVA} features provided by~\cite{Farha_2018_CVPR}, while for Assembly101 we utilized TSM~\cite{lin2019tsm} features provided by~\cite{sener2022assembly101}. Further implementation details of our model for these datasets are provided in the  supp.\ material.

\textbf{Evaluation Protocol and Metrics.}
We evaluate our method using the $\alpha$ and $\beta$ from the protocol defined in~\cite{Farha_2018_CVPR}. 
Specifically, 
we test our network for $\alpha \in \{0.2, 0.3\}$ and $\beta \in \{0.1, 0.2, 0.3, 0.5\}$, where $\alpha$ and $\beta$ denote the percentages of frames of a video that are used as observation and future prediction, respectively.     
We evaluate our approach in the stochastic and deterministic settings.  
For the deterministic setting, we report mean over classes (MoC) accuracy as in~\cite{Farha_2018_CVPR, Ke_2019_CVPR, gong2022future}. 
In the stochastic setting, we generate multiple predictions for the same observed frames. As proposed in~\cite{farha2019uaaa}, we report two metrics: mean and top-1 MoC across $M=25$ predictions, called Mean MoC and Top-1 MoC, respectively.

\begin{table}[t]
\centering
\caption{
Comparison to the state of the art for stochastic anticipation on Breakfast, Assembly101 and 50Salads. * Indicates that we trained UAAA on Assembly101. 
}
\vspace{-0.3cm}
\resizebox{0.8\linewidth}{!}{%
\setlength{\tabcolsep}{3.5pt}
\begin{tabular}{lll rrrr | rrrr }
\toprule
\multirow{2}{*}{Dataset} & \multirow{2}{*}{Metric} & \multirow{2}{*}{Method}  &
\multicolumn{4}{c}{$\beta$ ($\alpha=0.2$)} & \multicolumn{4}{c}{$\beta$ ($\alpha=0.3$)} \\
\cline{4-11}
& & & \textit{0.1} & \textit{0.2} & \textit{0.3} & \textit{0.5} & \textit{0.1} & \textit{0.2} & \textit{0.3} & \textit{0.5} \\
\midrule
\midrule
\multirow{4}{*}{Breakfast} & \multirow{2}{*}{Mean MoC} & Tri-gram~\cite{farha2019uaaa} 
& 15.4 & 13.7 & 12.9 & 11.9
& 19.3 & 16.6 & 15.8 & 13.9 \\

& & UAAA~\cite{farha2019uaaa}  
& 15.7 & 14.0 & 13.3 & 13.0  
& 19.1 & 17.2 & 17.4 & 15.0   \\
& & Ours 
& \textbf{24.0} & \textbf{22.0} & \textbf{21.4} & \textbf{20.6} 
& \textbf{29.1} & \textbf{26.8} & \textbf{25.3} & \textbf{24.2} \\

\cmidrule[0.8pt]{2-11}

& \multirow{2}{*}{Top-1 MoC} & UAAA~\cite{farha2019uaaa}                     
& 28.9 & 28.4 & 27.6 & 28.0 
& 32.4 & 31.6 & 32.8 & 30.8   \\
& & Ours                   
& \textbf{51.2} & \textbf{47.3} & \textbf{45.6} & \textbf{45.0}
& \textbf{54.0} & \textbf{50.4} & \textbf{49.6} & \textbf{47.8} \\
\midrule
\midrule

\multirow{5}{*}{Assembly101} & \multirow{3}{*}{Mean MoC} & Tri-gram 
& 2.8 & 2.2 & 1.9 & 1.5 
& 3.5 & 2.7 & 2.3 & 1.8 \\

& & UAAA*~\cite{farha2019uaaa} 
& 2.7 & 2.1 & 1.9 & 1.7 
& 2.4 & 2.1 & 1.9 & 1.7 \\

& & Ours 
& \textbf{6.4} & \textbf{4.4} & \textbf{3.5} & \textbf{2.8} 
& \textbf{5.9} & \textbf{4.2} & \textbf{3.5} & \textbf{2.9} \\

\cmidrule[0.8pt]{2-11}

& \multirow{2}{*}{Top-1 MoC} & UAAA*~\cite{farha2019uaaa}  
& 6.9 & 5.9 & 5.6 & 5.1 
& 5.9 & 5.5 & 5.2 & 4.9 \\

& & Ours 
& \textbf{18.0} & \textbf{12.8} & \textbf{9.9} & \textbf{7.7} 
& \textbf{16.0} & \textbf{11.9} & \textbf{10.2} & \textbf{7.7} \\

\midrule
\midrule

\multirow{5}{*}{50Salads} & \multirow{3}{*}{Mean MoC} & Tri-gram~\cite{farha2019uaaa}
& 21.4 & 16.4 & 13.3 & 9.4  
& 24.6 & 15.6 & 11.7 & 8.6 \\

& & UAAA~\cite{farha2019uaaa} 
& 23.6 & 19.5 & \textbf{18.0} & \textbf{12.8}
& 28.0 & 18.0 & \textbf{14.8} & \textbf{12.1} \\

& & Ours 
& \textbf{28.3} & \textbf{22.1} & 17.8 & 11.7
& \textbf{29.9} & \textbf{18.5} & 14.2 & 10.6 \\

\cmidrule[0.8pt]{2-11}

& \multirow{2}{*}{Top-1 MoC} & UAAA~\cite{farha2019uaaa}  
& 53.5 & 43.0 & 40.5 & \textbf{33.7} 
& 56.4 & 42.8 & 35.8 & 30.2 \\
& & Ours 
& \textbf{69.6} & \textbf{55.8} & \textbf{45.2} & 28.1
& \textbf{66.2} & \textbf{44.9} & \textbf{39.2} & \textbf{31.0} \\
\bottomrule
\end{tabular}}
\vspace{-0.4cm}
\label{tab:results_main_diffusion}
\end{table}

\vspace{-0.2cm}
\subsection{Stochastic Anticipation}
\vspace{-0.2cm}

\begin{table}[b!]
\vspace{-0.3cm}
\begin{minipage}{0.45\textwidth}
    \captionof{table}{
    Ablation on the num. of diffusion inference steps and stages in GTAN on Breakfast. Numbers show Mean MoC accuracy.
    }
    \vspace{-0.3cm}
    \centering
    \resizebox{1.0\linewidth}{!}{%
    \setlength{\tabcolsep}{3.5pt}
    \begin{tabular}{ll rrrr | rrrr}
    \toprule
    Num. & Num. & \multicolumn{4}{c}{$\beta$ ($\alpha=0.2$)} & \multicolumn{4}{c}{$\beta$ ($\alpha=0.3$)} \\
    \cline{3-10}
    stages & steps & \textit{0.1} & \textit{0.2} & \textit{0.3} & \textit{0.5} & \textit{0.1} & \textit{0.2} & \textit{0.3} & \textit{0.5} \\
    \midrule
    \midrule
    5 & 1 &
    19.6 & 17.7 & 17.3 & 15.9 & 
    24.9 & 22.6 & 22.5 & 20.0 \\
    
    5 & 10 &
    23.0  & 20.7 & 20.3 & 19.5 & 
    28.0 & 25.7 & 24.7 & 23.8 \\
    
    5 & 50 &
    \underline{24.0}& \textbf{22.0} & \textbf{21.4} & \textbf{20.6} 
    & \underline{29.1} & \textbf{26.8} & \textbf{25.3} & \textbf{24.2} \\
    
    5 & 100 & 
    \textbf{24.2} & 21.8 & 21.3 & 20.5
    & 29.1 & 26.7 & 25.0 & 24.0 \\    
    
    \midrule
    \midrule
    
    1 & 50 
    & 23.7 & 21.9 & 21.1 & 20.2
    & 29.2 & 26.7 & 25.3 & 23.9 \\
    
    3 & 50
    & 23.4 & 21.6 & 21.0 & 20.4 
    & \textbf{29.6} & 26.8 & 25.2 & \textbf{24.4}  \\
    
    5 & 50 
    & \textbf{24.0} & \textbf{22.0} & \textbf{21.4} & \textbf{20.6} 
    & \underline{29.1} & \textbf{26.8} & \textbf{25.5} & \underline{24.2} \\
    
    \midrule
    \midrule
    
    5 & 50 
    & \textbf{24.0} & \textbf{22.0} & \textbf{21.4} & \textbf{20.6}
    & 29.1 & 26.8 & 25.3 & \textbf{24.2} \\
    
    1 & 250 
    & 23.9 & 21.8 & 21.0 & 20.0
    & \textbf{29.6} & \textbf{27.0} & \textbf{25.5} & 23.9\\
    \bottomrule
    \end{tabular}
    }

    \label{tab:abl_diff_steps_stages_breakfast}
\end{minipage}
\hspace{0.2cm}
\begin{minipage}{0.5\textwidth}
    \caption{
    Ablation on GTAN architecture on Breakfast. Numbers show Mean MoC accuracy.
    }
    \vspace{-0.3cm}
    \centering
    \resizebox{1.0\linewidth}{!}{%
    \setlength{\tabcolsep}{3.5pt}
    \begin{tabular}{l rrrr | rrrr}
    \toprule
    \multirow{2}{*}{Method}  & \multicolumn{4}{c}{$\beta$ ($\alpha=0.2$)} & \multicolumn{4}{c}{$\beta$ ($\alpha=0.3$)} \\
    \cline{2-9}
    & \textit{0.1} & \textit{0.2} & \textit{0.3} & \textit{0.5} & \textit{0.1} & \textit{0.2} & \textit{0.3} & \textit{0.5} \\
    
    \midrule
    \midrule
    
    Ours
    & \textbf{24.0} & \textbf{22.0} & \textbf{21.4} & \textbf{20.6}  
    & \textbf{29.1} & \textbf{26.8} & \textbf{25.3} & \textbf{24.2}   \\

    \midrule
    \midrule
    Ours w/o GC
    & 23.0 & 21.1 & 20.6 & 19.8 
    & 27.6 & 25.7 & 24.3 & 23.5 \\

    Ours w/o Dil. GC
    & 22.9 & 21.1 & 20.6 & 20.2 
    & 28.6 & 26.0 & 24.7 & 23.9 \\

    Aslan~\etal~\cite{aslan2022gtcn}
    & 18.8 & 17.3 & 16.7 & 15.7 
    & 20.7 & 18.9 & 18.8 & 16.7 \\

    \midrule 
    
    Part.\ Conv.~\cite{Liu2018ImageIF} 
    & 23.2 & 21.5 & 20.9 & 20.1 
    & 27.9 & 25.6 & 24.4 & 23.6 \\
    
    SE~\cite{Hu2017SqueezeandExcitationN} 
    & 22.4 & 21.0 & 20.4 & 19.9
    & 28.0 & 25.7 & 24.5 & 23.6 \\
  
    \bottomrule
\end{tabular}
}
\label{tab:abl_diffusion_arch_breakfast}
\end{minipage}
\vspace{-0.5cm}
\end{table}

\textbf{Comparison with State of the Art}. We first compare our proposed stochastic approach with the state of the art for stochastic anticipation on the Breakfast, Assembly101 and 50Salads datasets. In Sec.~\ref{sec:exp:deter}, we compare our approach to deterministic approaches. The comparison of our diffusion-based model with UAAA~\cite{farha2019uaaa} and tri-gram baseline~\cite{farha2019uaaa} is presented in Tab.~\ref{tab:results_main_diffusion}. UAAA~\cite{farha2019uaaa} and the tri-gram baseline are the only available probabilistic models with a comparable evaluation protocol since~\cite{zhao2020async} only uses ground-truth action labels as their observations. 
\cite{farha2019uaaa} is a two-step approach that first predicts action labels for the observed frames and then forecasts the future actions. 
To compare our method to~\cite{farha2019uaaa} on the Assembly101 dataset, we used MS-TCN~\cite{AbuFarha2019MSTCNMT} for the first step. We trained a MS-TCN network for each value of $\alpha$ using full supervision, but only the observed frames as training data, \ie, the first 20\% or 30\%, respectively. We then trained UAAA~\cite{farha2019uaaa}. Note that two-step approaches~\cite{farha2019uaaa,zhao2020async} do not model any uncertainty in the observation.    
As shown in Tab.~\ref{tab:results_main_diffusion}, our method outperforms~\cite{farha2019uaaa} with a large margin on Breakfast and Assembly101 at both Mean and Top-1 MoC accuracy with improvements across all observation and anticipation ratios. On the 50Salads dataset, our approach outperforms~\cite{farha2019uaaa} at Top-1 MoC, while the Mean MoC accuracy is on par.

In Fig.~\ref{fig:qualitative_res_pron}, we present predicted samples from our model. In the first example, the action sequence involves the high-level activity `Scrambled Eggs'. However, the observed part of the sequence contains only the action `Butter Pan', shared with another high-level class, `Fry Eggs'. Thus, based solely on this segment, distinguishing the underlying activity is challenging. Our network predicts sequences belonging to either `Scrambled Eggs' or `Fry Eggs', demonstrating ambiguity in order, length, and presence/absence of actions within categories.
In the second example, the ground-truth activity is `Cereals'. Poor lighting conditions make recognizing actions in the observed segment difficult and ambiguous. Consequently, our model classifies observed actions differently across samples, leading to consistent yet different predictions. By addressing uncertainty in observations, our model produces correct predictions despite the poor quality of the observed segment.
We provide more qualitative results in the supp.\ material.

\label{sec:ms_gtcn}

\subsection{Ablation Study}
\vspace{-0.2cm}

\begin{figure}[b!]
    \centering
    \vspace{-0.4cm}
    \includegraphics[width=0.9\linewidth]{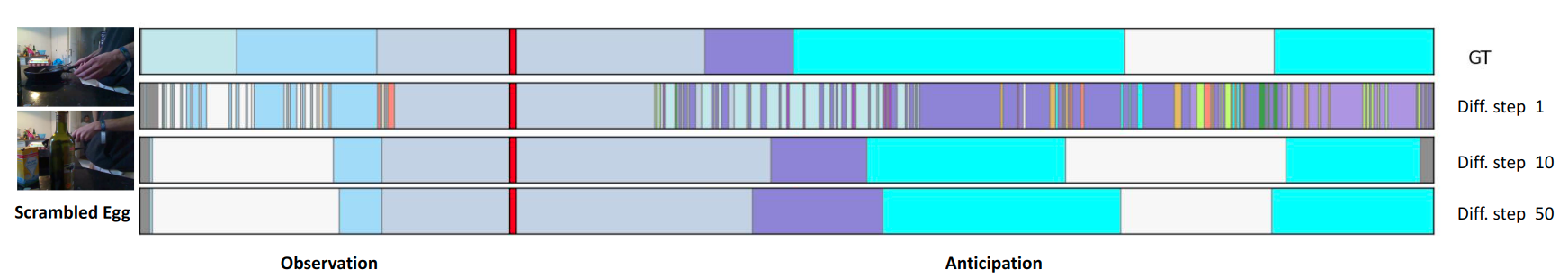}
    \vspace{-0.2cm}
    \caption{Qualitative results of our proposed GTD with different numbers of inference diffusion steps. Best viewed zoomed in.}
    \label{fig:diff_step}
\end{figure}

\textbf{Number of Inference Steps and Stages.}
In Tab.~\ref{tab:abl_diff_steps_stages_breakfast}, we explored the impact of varying the number of denoising steps $D$ in our diffusion model while keeping the number of GTAN stages fixed. Increasing the number of steps from 10 to 50 led to improved accuracy, but further increasing it to 100 did not yield additional benefits. Fig.~\ref{fig:diff_step} illustrates how predictions evolve with different denoising steps: noise decreases significantly after 10 steps, with further refinement observed with 50 steps.
Next, with $D$ fixed at 50, we evaluated the influence of different numbers of stages. Smaller networks (1 and 3 stages) exhibited lower accuracy with 50 denoising steps, but with 250 steps, the single-stage GTAN's performance approached that of 5 stages. For all other experiments, we use 5 stages and 50 denoising steps.  

\begin{table}[t!]
\begin{minipage}{0.55\textwidth}
    \centering
    \vspace{0.1cm}
    \includegraphics[scale=0.125]{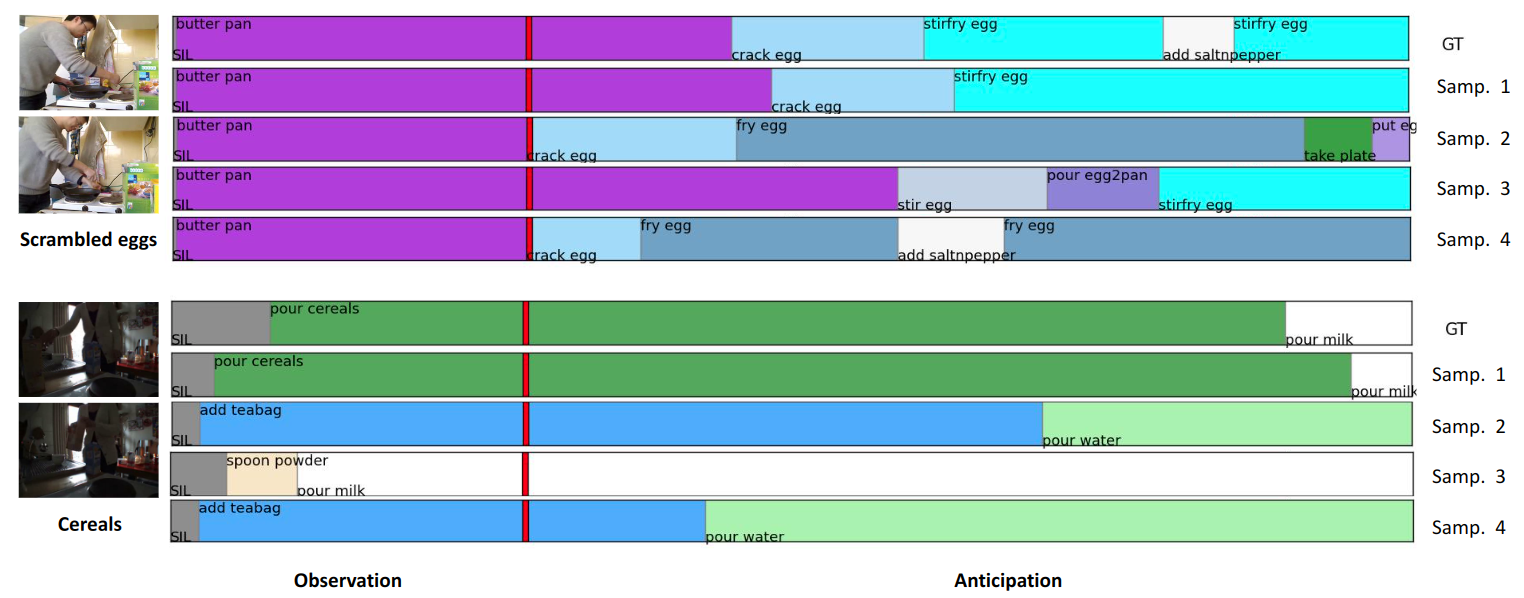}
    \captionof{figure}{Qualitative results of our proposed GTD for stochastic long-term action anticipation on Breakfast. Best viewed zoomed in. }
    \label{fig:qualitative_res_pron}
\end{minipage}
\hspace{0.2cm}
\begin{minipage}{0.4\textwidth}
    \centering
    \includegraphics[scale=0.14]{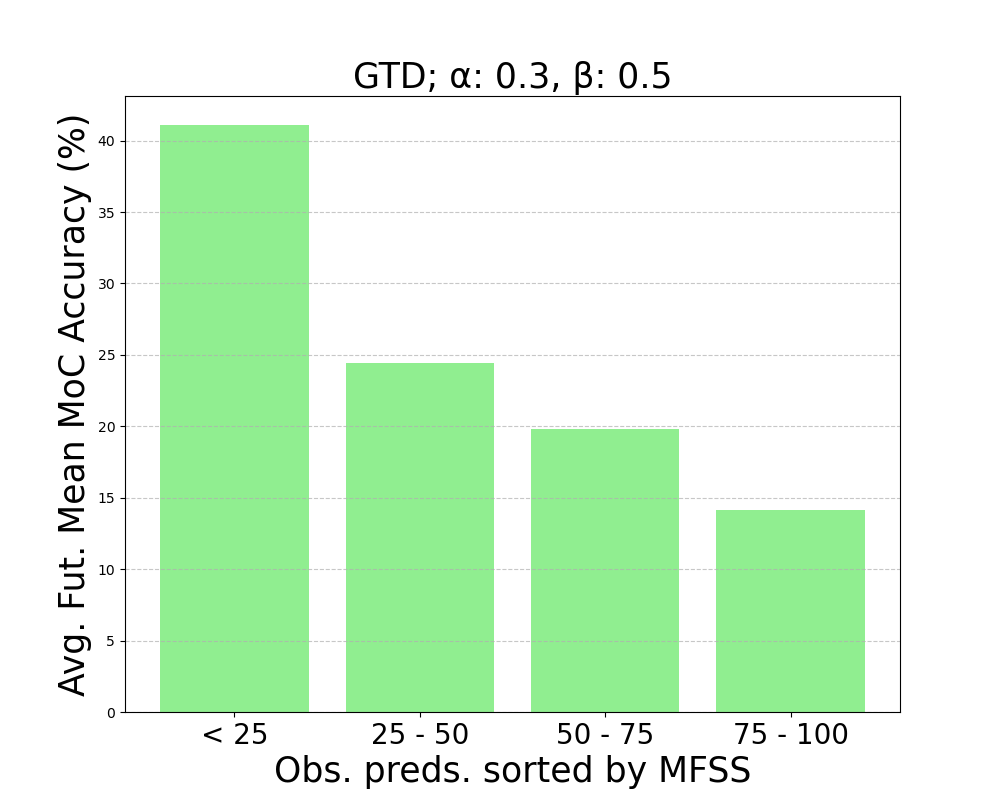}
    \captionof{figure}{Mean MoC of GTD for sequences sorted by MFSS diversity for the obs. part on Breakfast.}
    \label{fig:diff_moc_mfss}
\end{minipage}
\vspace{-0.9cm}
\end{table}

\textbf{Gated Convolution.}
In Tab.~\ref{tab:abl_diffusion_arch_breakfast}, we assess the impact of gated convolution on our GTAN generator. Removing the gate convolution branch (see Fig.~\ref{fig:arch}) from the GTA block and leaving only the feature convolution (Ours w/o GC) leads to lower performance, highlighting the necessity of the gating mechanism. Similarly, removing the dilation factor from the gate convolution branch (Ours w/o Dil. GC) also decreases performance, as expected. Substituting our GTAN architecture with the gated temporal convolutional network proposed by Aslan~\etal\cite{aslan2022gtcn} results in a significant decline in performance, indicating its unsuitability for dense anticipation.

We also explore alternatives to gated convolutions, including manual mask adaptation and channel-wise feature reweighing, by replacing gated convolutions with partial convolution~\cite{Liu2018ImageIF} and squeeze-and-excitation~\cite{Hu2017SqueezeandExcitationN} blocks, respectively. Although manual mask adaptation outperforms the model without gating, it falls short compared to our proposed approach, highlighting the advantage of learning to perform gating in a data-driven manner. Employing the squeeze-and-excitation block instead of gated convolutions yields better performance compared to previous alternatives, but it remains inferior to our proposed method. This emphasizes the importance of the temporal component of gated convolutions. We provide further ablations of the GTAN design and a qualitative example of learned gates in the supp.\ material.

\begin{table}[t!]
\begin{minipage}{0.5\linewidth}
    \caption{
    Ablation on modelling observation uncertainty (o.u.) on Breakfast. 
    }
    \vspace{-0.3cm}
    \centering
    \resizebox{1.0\linewidth}{!}{%
    \setlength{\tabcolsep}{3.5pt}
    \begin{tabular}{ll rrrr | rrrr}
    \toprule
     & & \multicolumn{4}{c}{$\beta$ ($\alpha=0.2$)} & \multicolumn{4}{c}{$\beta$ ($\alpha=0.3$)} \\
    \cline{3-10}
    Metric & Method & \textit{0.1} & \textit{0.2} & \textit{0.3} & \textit{0.5} & \textit{0.1} & \textit{0.2} & \textit{0.3} & \textit{0.5} \\
    \midrule
    \midrule
    
    \multirow{2}{*}{Mean MoC} & Ours 
    & \textbf{24.0} & \textbf{22.0} & \textbf{21.4} & \textbf{20.6}
    & 29.1 & 26.8 & \textbf{25.3} & \textbf{24.2} \\
    
    &  Ours w/o o.u.
    & 23.7 & 21.4 & 20.5 & 19.9
    & \textbf{29.8} & \textbf{27.1} & \textbf{25.3} & 24.1 \\
    
    \midrule 
    
    \multirow{2}{*}{Top-1 MoC} & Ours
    & \textbf{51.2} & \textbf{47.3} & \textbf{45.6} & \textbf{45.0}
    & \textbf{54.0} & \textbf{50.4} & \textbf{49.6} & \textbf{47.8} \\
    
    &  Ours w/o o.u.
    & 42.8 & 40.7 & 38.8 & 38.4
    & 48.4 & 44.2 & 44.9 & 43.5 \\

    \midrule
    
    \multirow{2}{*}{MFSS} & Ours
    & \textbf{41.5} &  \textbf{44.3} & \textbf{45.6} & \textbf{48.4}
    & \textbf{33.7} & \textbf{36.6} & \textbf{38.0} & \textbf{41.6} \\
    
    &  Ours w/o o.u.
    & 31.2 & 34.7 & 36.3 & 39.4
    & 26.1 & 29.4 & 30.8 & 34.3 \\
    \bottomrule
    
    \end{tabular}
    }
    \label{tab:abl_diff_obs_unc}
\end{minipage}
\hspace{0.2cm}
\begin{minipage}{0.45\linewidth}
    \vspace{-0.1cm}
    \caption{
    Ablation on the diffusion loss type on Breakfast. 
    }
    \vspace{-0.3cm}
    \centering
    \resizebox{1.0\linewidth}{!}{%
    \setlength{\tabcolsep}{3.5pt}
    \begin{tabular}{ll rrrr | rrrr}
    \toprule
     & & \multicolumn{4}{c}{$\beta$ ($\alpha=0.2$)} & \multicolumn{4}{c}{$\beta$ ($\alpha=0.3$)} \\
    \cline{3-10}
    Metric & Loss & \textit{0.1} & \textit{0.2} & \textit{0.3} & \textit{0.5} & \textit{0.1} & \textit{0.2} & \textit{0.3} & \textit{0.5} \\
    \midrule
    \midrule
    
    \multirow{3}{*}{Mean MoC} & MSE 
    & 24.0 & 22.0 & 21.4 & 20.6 
    & 29.1 & 26.8 & 25.3 & 24.2 \\

    & CE &  
    \textbf{26.2} & \textbf{23.3}  & \textbf{23.9} & \textbf{22.6} & 
    \textbf{32.1} & \textbf{29.2} & \textbf{27.5} & \textbf{26.3}  \\
    
    \midrule
    
    \multirow{3}{*}{Top-1 MoC} & MSE 
    & \textbf{51.2} & \textbf{47.3} & \textbf{45.6} & \textbf{45.0}
    & \textbf{54.0} & \textbf{50.4} & \textbf{49.6} & \textbf{47.8} \\

    & CE 
    & 46.9 & 42.2 & 41.7 & 40.7
    & 48.8 & 46.5 & 45.3 & 44.0 \\

    \midrule

    \multirow{3}{*}{MFSS} & MSE 
    & \textbf{41.5} & \textbf{44.3} & \textbf{45.6} & \textbf{48.4}
    & \textbf{33.7} & \textbf{36.6} & \textbf{38.0} & \textbf{41.6} \\

    & CE 
    & 29.0 & 31.8 & 33.5 & 35.0
    & 24.0 & 26.9 & 28.2 & 30.0 \\

    \bottomrule
    
    \end{tabular}
    }
    \label{tab:abl_diff_loss_type_breakfast}
\end{minipage}
\vspace{-0.6cm}
\end{table}

\textbf{Modelling Observation Uncertainty.} We conduct an experiment to investigate the impact of modelling observation ambiguity on the performance of our model. To this end, we utilize an MS-TCN~\cite{AbuFarha2019MSTCNMT} to predict labels of the observed part, which serves as the condition vector for GTD instead of the frame features $F$. Also, we compute the loss function $L_{stoch}$ only for the future frames.
The results of this experiment are shown in Tab.~\ref{tab:abl_diff_obs_unc}. While the Mean MoC of both methods remains comparable, the Top-1 MoC of the model without uncertainty modelling drops significantly, indicating its limited ability to generate diverse predictions.
To directly measure the prediction diversity, we additionally introduce a new metric - Mean Framewise Sample Similarity (MFSS). MFSS calculates diversity as the mean normalized pairwise sample distance averaged over the videos in the split:
\vspace{-0.1cm}
\begin{align}
    \text{\footnotesize$MFSS$} = \text{\footnotesize$\frac{1}{Z}\sum_{z=1}^{Z}\left(\frac{2}{M(M-1)}\sum_{1 \leq i < j \leq M} 100\left(1 - \frac{1}{N}\sum_{n=1}^{N} \mathbf{1}(\hat{Y}_{z,i}^n = \hat{Y}_{z, j}^n) \right)\right)$},
\end{align}
where $Z$ is the total number of videos.
As evident from the results, omitting the observation uncertainty modelling leads to a reduction in the diversity of predictions.
In Fig.~\ref{fig:diff_moc_mfss}, we demonstrate the correlation between Mean MoC anticipation accuracy and uncertainty of the GTD model for the observation.
To this end, we sorted the videos into four groups based on the MFSS of the predictions on the observed parts of the videos and calculated the average Mean MoC for anticipation for each group. For sequences with high MFSS, GTD is uncertain about the observed actions.    
The plot reveals an inverse correlation between future Mean MoC and observed MFSS since future actions are harder to predict for videos with higher observation uncertainty. 
This analysis underscores the significance of modelling ambiguity in observed frames for future action prediction.

\textbf{Diffusion Loss Type.}
We employ Mean Squared Error (MSE) as the loss function for our diffusion approach, following prior work~\cite{ho2020denoisingDiff, chen2023analog}. Additionally, we evaluate the impact of using the Cross Entropy (CE) loss and present the results in Tab.~\ref{tab:abl_diff_loss_type_breakfast}. Notably, the model trained with MSE loss achieves the highest Top-1 MoC accuracy and MFSS, while the model trained with CE loss attains the best Mean MoC.
These results of the CE-trained network can be attributed to additional Softmax normalization required by this loss. The Softmax restricts the handling of uncertainty in early diffusion steps, emphasizing the most likely action class prematurely. 
Directly measuring diversity using the MFSS metric confirms these observations, with the MSE-trained model exhibiting higher diversity than the CE-trained model. We used the MSE loss in our experiments.

\begin{table}[t!]
\centering
\caption{
Comparison with state-of-the-art methods for deterministic anticipation on the Breakfast dataset. * Indicates retrained results.
}
\vspace{-0.3cm}
\resizebox{0.6\linewidth}{!}{%
\setlength{\tabcolsep}{3.5pt}
\begin{tabular}{ll rrrr | rrrr}
\toprule
\multirow{2}{*}{Dataset} & \multirow{2}{*}{Method}  & \multicolumn{4}{c}{$\beta$ ($\alpha=0.2$)} & \multicolumn{4}{c}{$\beta$ ($\alpha=0.3$)} \\
\cline{3-10}
& & \textit{0.1} & \textit{0.2} & \textit{0.3} & \textit{0.5} & \textit{0.1} & \textit{0.2} & \textit{0.3} & \textit{0.5} \\
\midrule
\midrule

\multirow{8}{*}{Breakfast} & RNN~\cite{Farha_2018_CVPR}
& 18.1 & 17.2 & 15.9 & 15.8
& 21.6 & 20.0 & 19.7 & 19.2 \\

& CNN~\cite{Farha_2018_CVPR} 
& 17.9 & 16.3 & 15.4 & 14.5
& 22.4 & 20.1 & 19.7 & 18.8 \\

& UAAA (mode)~\cite{farha2019uaaa}
& 16.7 & 15.4 & 14.5 & 14.2
& 20.7 & 18.3 & 18.4 & 16.9 \\

& Time Cond.~\cite{Ke_2019_CVPR}
& 18.4 & 17.2 & 16.4 & 15.8
& 22.8 & 20.4 & 19.6 & 19.8 \\

& TempAgg~\cite{sener2020temporal} 
& 24.2 & 21.1 & 20.0 & 18.1
& 30.4 & 26.3 & 23.8 & 21.2 \\

& Cycle Cons.~\cite{farha2020gcpr} 
& 25.9 & 23.4 & 22.4 & 21.5
& 29.7 & 27.4 & 25.6 & 25.2 \\

& FUTR~\cite{gong2022future} 
& 27.7 & 24.6 & 22.8 & 22.1 
& 32.3 & 29.9 & 27.5 & 25.9 \\

& \textbf{Ours}                      
& \textbf{28.8} & \textbf{26.3} & \textbf{25.8} & \textbf{26.0}
& \textbf{35.5} & \textbf{32.9} & \textbf{30.5} & \textbf{29.6} \\

\midrule
\midrule
\multirow{3}{*}{Assembly101}& UAAA (mode)*~\cite{farha2019uaaa} & 2.7 & 2.1 & 1.8 & 1.6 & 2.4 & 2.1 & 1.9 & 1.7 \\
& FUTR*~\cite{gong2022future} & 7.5 & 5.5 & 4.7 & 4.1 & 7.8 & 6.0 & 5.2 & 4.0 \\
& Ours                       & \textbf{9.0} & \textbf{6.8} & \textbf{6.6} & \textbf{5.5} & \textbf{8.4} & \textbf{6.8} & \textbf{6.0} & \textbf{5.0} \\

\bottomrule
\end{tabular}
}
\label{tab:results_main_determ}
\vspace{-0.3cm}
\end{table}

\subsection{Deterministic Anticipation}\label{sec:exp:deter}
\vspace{-0.2cm}

\textbf{Comparison with State of the Art}.
The majority of dense long-term anticipation methods~\cite{gong2022future, farha2020gcpr, sener2020temporal, Ke_2019_CVPR, Farha_2018_CVPR} has been trained and evaluated in the deterministic setting and we compare with these methods in Tab.~\ref{tab:results_main_determ}. We do not compare with Anticipatr~\cite{nawhal2022anticipatr} since it uses a different protocol~\cite{zhong2023survey}. We also compare with UAAA~\cite{farha2019uaaa} in the deterministic setting (mode). On Assembly 101, we compare with UAAA (mode) and FUTR~\cite{gong2022future} using the publicly available code for training and evaluation. We provide results for the 50Salads dataset, qualitative comparisons and ablation studies for the deterministic GTAN in the supp.\ material.
On both Breakfast and Assembly101, our method outperforms all methods that use the same evaluation protocols. Compared to the previously best-performing method FUTR~\cite{gong2022future}, our method shows a substantial improvement in particular on Breakfast for long-term prediction ($\beta=0.5$) where MoC is increased by +3.9 and +3.7 for $\alpha=0.2$ and $\alpha=0.3$, respectively.     

\vspace{-0.2cm}
\section{Conclusion}
\vspace{-0.3cm}
We have proposed a Gated Temporal Diffusion network to address the task of stochastic long-term dense action
anticipation. As the backbone for our diffusion model, we introduced a Gated Anticipation Network (GTAN) that allows for mutual modelling of the actions in the observed and future frames. In this way, the uncertainty is not only modelled for the future but also for the observed part. We demonstrated that the approach generates different predictions for the observed frames in case of ambiguities due to poor light conditions and that these ambiguities impact the future predictions. Furthermore, we demonstrated that GTAN can be applied to deterministic long-term dense action anticipation as well. In our experiments, we showed that our model achieves state-of-the-art results on three datasets in deterministic and stochastic settings. A limitation of our proposed model is its current efficiency. For example, the average time of generating a single prediction on the Breakfast dataset, which has an average anticipation horizon of 1.15 minutes, is 3.8 seconds. While this is sufficient for mid-term action planning, i.e.\ range of minutes, a further reduction of inference time is needed. This can be achieved by using techniques to accelerate inference of diffusion models like distillation~\cite{kohler2024imagine} or DeepCache~\cite{ma2024deepcache}.

\section*{Acknowledgements}
The work has been supported by the Deutsche Forschungsgemeinschaft (DFG, 
German Research Foundation) GA 1927/4-2 (FOR 2535 Anticipating Human 
Behavior), by the Federal Ministry of Education and Research (BMBF) 
under grant no. 01IS22094E WEST-AI, the project
iBehave (receiving funding from the programme “Netzwerke 2021”, an 
initiative of the Ministry of Culture and Science of the State of 
Northrhine Westphalia), and the ERC Consolidator Grant FORHUE 
(101044724). The authors gratefully acknowledge the Gauss Centre for 
Supercomputing e.V. (www.gauss-centre.eu) for funding this project by 
providing computing time through the John von Neumann Institute for 
Computing (NIC) on the GCS Supercomputer JUWELS at Jülich Supercomputing 
Centre (JSC). The sole responsibility for the content of this 
publication lies with the authors.

%
%
\bibliographystyle{splncs04}
\bibliography{main}


\title{Supplementary Material: Gated Temporal Diffusion for Stochastic Long-Term Dense Anticipation} 

\titlerunning{Gated Temporal Diffusion for Stochastic Long-Term Dense Anticipation}

\author{Olga Zatsarynna\inst{1}*\and
Emad Bahrami\inst{1}*\and
Yazan Abu Farha\inst{2} \and 
Gianpiero Francesca\inst{3} \and
Juergen Gall\inst{1,4}}

\authorrunning{O. Zatsarynna et al.}


\institute{University of Bonn, Germany \\
\and
Birzeit University, Palestine\\
\and
Toyota Motor Europe, Belgium \\
\and
Lamarr Institute for Machine Learning and Artificial Intelligence, Germany \\ *Equal contribution 
}

\authorrunning{O. Zatsarynna et al.}

\maketitle

In Sec.~\ref{sec:impl}, we provide implementation details.
In Sec.~\ref{sec:stoch}, we provide additional ablation studies and qualitative results for stochastic long-term anticipation. In Sec.~\ref{sec:determ}, we present additional results for deterministic long-term anticipation.

\section{Implementation Details}
\label{sec:impl}
We implemented our models in Pytorch. For both stochastic and deterministic settings, we used $5$ stages and $9$ layers for the GTAN model. For the filters of the gated convolutions, we set the kernel size to $3$, the number of channels to $64$, and the dilation factor to $2^l$, \ie, the dilation factor is increased by a factor of two with each layer $l$ in a stage.
We trained our network with a batch size of $16$ using the Adam~\cite{kingma2014adam} optimizer with the learning rate $5 \cdot 10^{-4}$ on the Breakfast and Assembly101 datasets. For the 50salads dataset, we used the batch size of $8$, set the learning rate to $10^{-3}$.

For stochastic anticipation, we trained our GTD model with $T=1000$ denoising steps. During inference, we applied DDIM~\cite{song2021denoising} sampling procedure and used a subset of $D=50$ denoising steps for generating predictions on the Breakfast and Assembly101 datasets and $D=10$ steps on the 50Salads dataset.

\section{Stochastic Long-Term Anticipation}
\label{sec:stoch}

\subsection{Ablation Study}
\label{sec:stoch_abl}

\subsubsection{Gated Convolution.}
\begin{figure}[t!]
    \centering
    \includegraphics[scale=0.25]{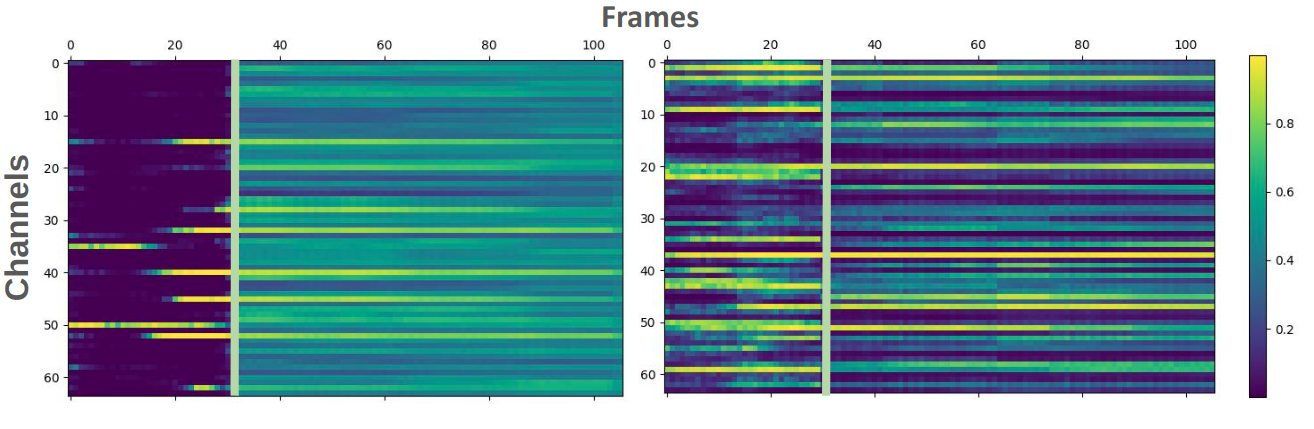}
    \vspace{-0.3cm}
    \caption{Visualization of gates from the GTAN. Both outputs are taken from the second stage from layers $l=2$ (left) and $l=8$ (right). The vertical green line marks the boundary between the observed and future frames. }
    \label{fig:gates}
\end{figure}

In Fig.~\ref{fig:gates}, we illustrate two examples of gate values taken from two distinct layers of the second stage of our proposed network. In both cases, there is a clear difference between observed and future frames visible, not just in the early layers of the network.
For instance, at layer 8, there are some channels where mainly the frames from the observation (channel 9) or mainly the future frames (channel 45) are considered.   
Additionally, the amount of mixing between observed and future frames changes for different layers. Starting from almost no mixing in the early layers, more and more mixing between unmasked and masked parts occurs as the network progresses.

\subsubsection{Modelling Observation Uncertainty.} 
\begin{figure}[h!]
    \centering
    \includegraphics[scale=0.17]{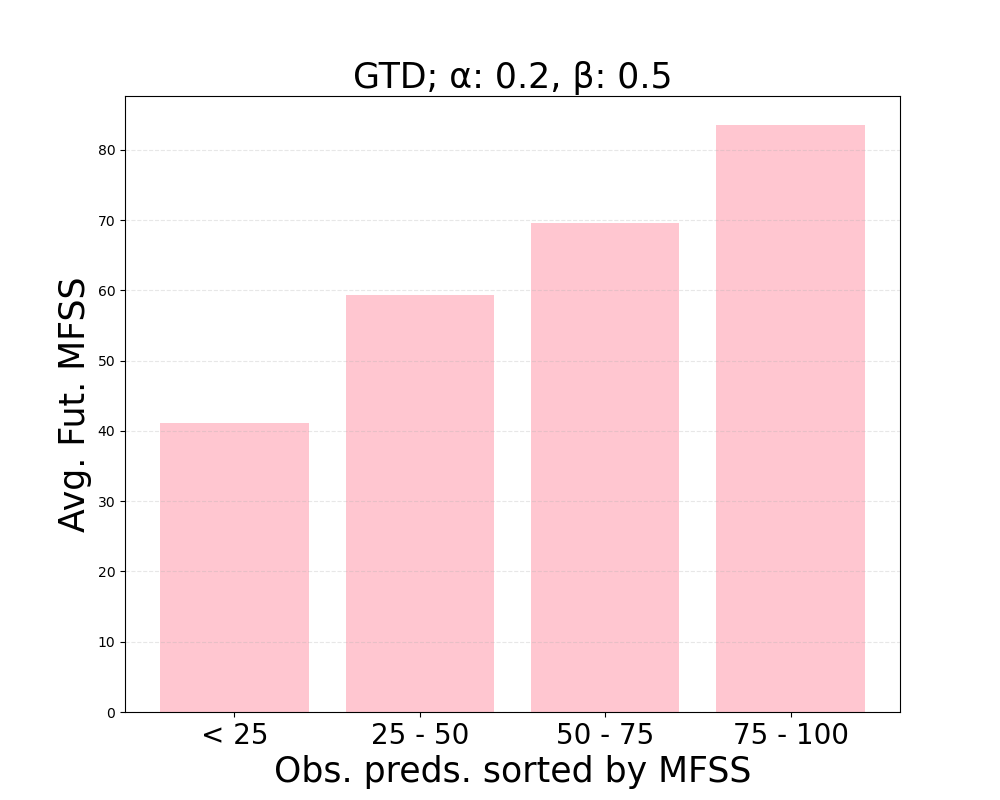}
    \includegraphics[scale=0.17]{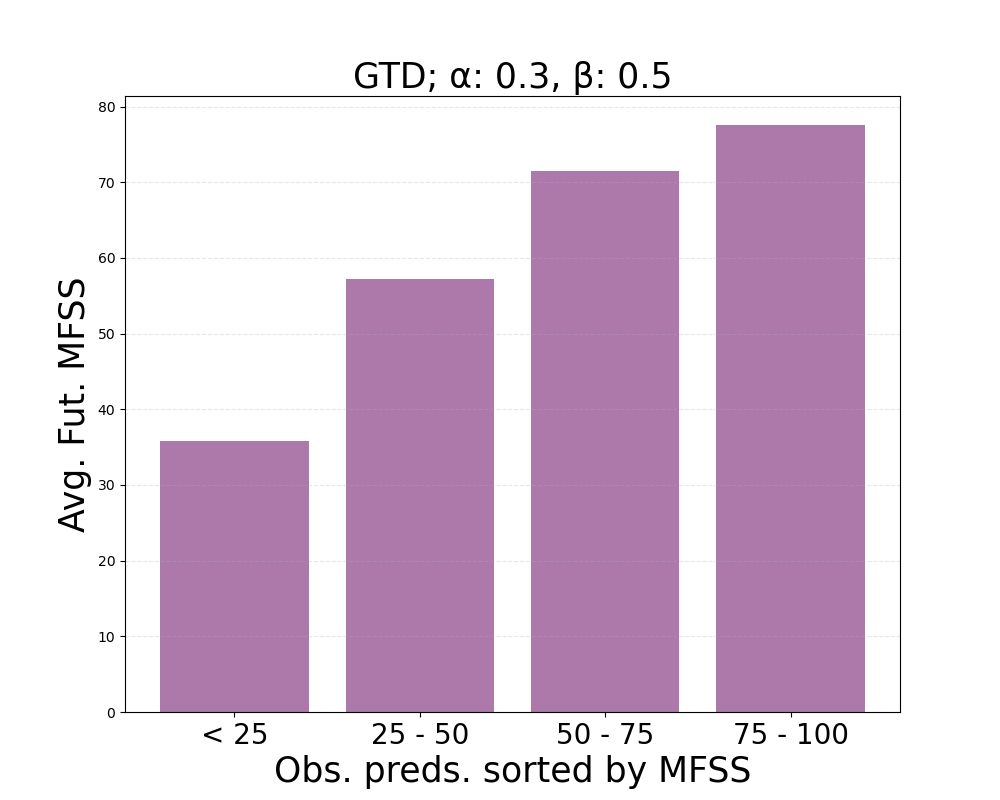}
    \vspace{-0.2cm}
    \caption{MFSS of GTD future predictions for sequences sorted by MFSS diversity for the observation part on Breakfast.}
    \label{fig:obs_mfss_fut_mfss}
\end{figure}

In Fig.~5 of the main paper, we showed the relationship between the future Mean MoC accuracy and the MFSS for the observed frames. Both metrics are computed over 25 predictions per observed video as shown in Figs.~\ref{fig:all_1}-\ref{fig:all_5}.
Additionally, in Fig.~\ref{fig:obs_mfss_fut_mfss}, we further visualize the relation between the MFSS of the predicted past and future actions. The grouping based on quartiles is the same as in Fig.~5 of the main paper. The first bar (first quartile) includes the videos with the lowest MFSS for the predicted actions of the observed part and the last bar (fourth quartile) includes the videos with the highest MFSS. For the videos of each quartile, we calculated the average MFSS for the future predictions, whereas Fig.~5 of the main paper plots the Mean MoC accuracy for the future predictions for each quartile.   
The plot demonstrates a direct correlation between the MFSS of the observed and future parts, as expected. When there is more uncertainty about the actions in the observed frames, the future predictions become more diverse.

\subsubsection{Effect of the number of samples.} In Tab.~\ref{tab:abl_diff_num_sampl_mean}
and Tab.~\ref{tab:abl_diff_num_sample_top} we demonstrate how varying the number of samples affects the Mean MoC and Top-1 MoC accuracy of our proposed GTD model.
On the one hand, the Top-1 MoC increases consistently, showing that our model generates diverse predictions. On the other hand, we observe a slight reduction in Mean MoC accuracy with the increased number of samples. This is expected due to the average value taken over increasingly diverse predictions.

\begin{table}[]
\begin{minipage}{0.48\linewidth}
    \caption{
    Effect of the number of samples on Mean MoC accuracy on Breakfast.
    }
    \vspace{-0.3cm}
    \centering
    \resizebox{1.0\linewidth}{!}{%
    \setlength{\tabcolsep}{3.pt}
    \begin{tabular}{l rrrr | rrrr}
    \toprule
     Num. & \multicolumn{4}{c}{$\beta$ ($\alpha=0.2$)} & \multicolumn{4}{c}{$\beta$ ($\alpha=0.3$)} \\
    \cline{2-9}
    samples & \textit{0.1} & \textit{0.2} & \textit{0.3} & \textit{0.5} & \textit{0.1} & \textit{0.2} & \textit{0.3} & \textit{0.5} \\
    \midrule
    \midrule
    
    5 & \textbf{24.2} & 21.9 & \textbf{21.7} & \textbf{20.9} 
                               & 29.1 & \textbf{26.9} & 25.4 & \textbf{24.7} \\
    
    10 & 23.9 & 21.8 & 21.5 & 20.9 
       & 29.1 & 26.8 & \textbf{25.5} & 24.4 \\
    
    20 & 24.0 & 21.9 & 21.4 & 20.7 
       & 29.2 & 26.7 & 25.3 & 24.2 \\
       
    25 & 24.0 & \textbf{22.0} & 21.4 & 20.6 
       & 29.1 & 26.8 & 25.3 & 24.2 \\
    \bottomrule 
    \end{tabular}
    }
    \label{tab:abl_diff_num_sampl_mean}
\end{minipage}
\hspace{0.2cm}
\begin{minipage}{0.48\linewidth}
    \caption{
    Effect of the number of samples on Top-1 MoC accuracy on Breakfast.
    }
    \vspace{-0.3cm}
    \centering
    \resizebox{1.0\linewidth}{!}{%
    \setlength{\tabcolsep}{3.pt}
    \begin{tabular}{l rrrr | rrrr}
    \toprule
     Num. & \multicolumn{4}{c}{$\beta$ ($\alpha=0.2$)} & \multicolumn{4}{c}{$\beta$ ($\alpha=0.3$)} \\
    \cline{2-9}
    samples & \textit{0.1} & \textit{0.2} & \textit{0.3} & \textit{0.5} & \textit{0.1} & \textit{0.2} & \textit{0.3} & \textit{0.5} \\
    \midrule
    \midrule
    
    5 & 38.8 & 36.2 & 35.4 & 34.2 
      & 41.9 & 39.2 & 37.7 & 36.5 \\

    10 & 44.0 & 40.7 & 39.8 & 38.9 
       & 47.1 & 44.2 & 43.5 & 42.2 \\

    20 & 49.2 & 45.5 & 44.2 & 43.7 
       & 52.3 & 49.0 & 47.9 & 46.3 \\

    25 & \textbf{51.2} & \textbf{47.3} & \textbf{45.6} & \textbf{45.0}
       & \textbf{54.0} & \textbf{50.4} & \textbf{49.6} & \textbf{47.8} \\
    \bottomrule
    
    \end{tabular}
    }
    \label{tab:abl_diff_num_sample_top}
\end{minipage}
\vspace{-0.1cm}
\end{table}

\subsubsection{Diffusion Loss Type.}
\begin{figure*}[t!]
    \centering
    \includegraphics[width=1.0\linewidth]{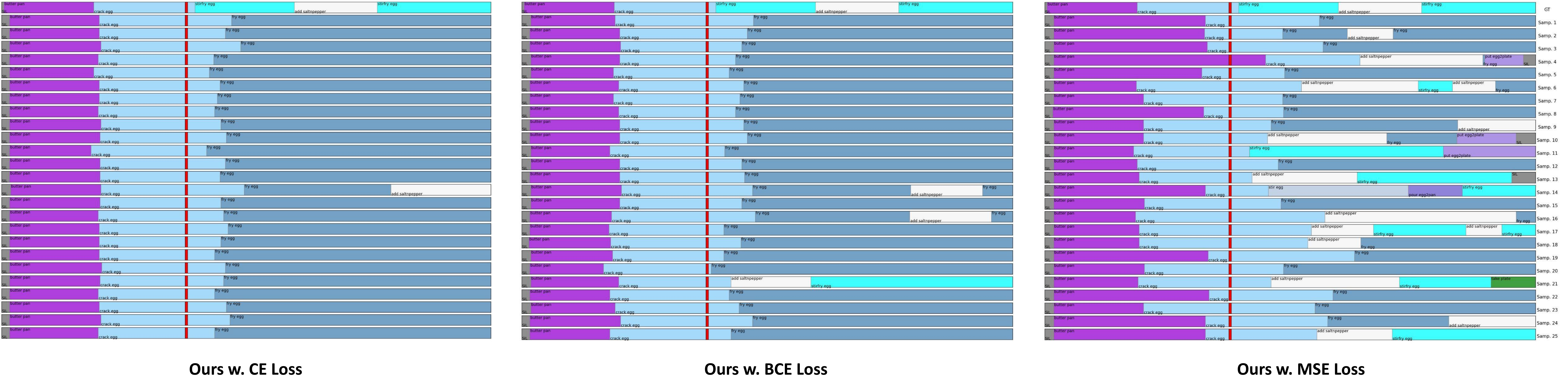}
    \vspace{-0.3cm}
    \caption{Qualitative comparison of the GTD model trained with three different loss terms. The example is from the Breakfast dataset. Best viewed zoomed in.}
    \label{fig:stoch_abl_diff_loss_type}
    \vspace{-0.2cm}
\end{figure*}

In Tab.~5 of the main paper, we compared the mean squared error (MSE) with the cross-entropy (CE) loss. In Tab.~\ref{tab:stoch_abl_diff_loss_type_breakfast}, we include results for the binary cross-entropy loss (BCE) as well. The BCE loss is between MSE and CE in terms of all metrics. The sigmoid of BCE results in a higher diversity compared to the softmax used by CE, but the diversity is still lower than for MSE. This is also visible in the example shown in Fig.~\ref{fig:stoch_abl_diff_loss_type}.

\begin{table}[t!]
    \caption{
    Ablation on the diffusion loss type on Breakfast. 
    }
    \vspace{-0.3cm}
    \centering
    \resizebox{0.7\linewidth}{!}{%
    \setlength{\tabcolsep}{3.5pt}
    \begin{tabular}{ll rrrr | rrrr}
    \toprule
     & & \multicolumn{4}{c}{$\beta$ ($\alpha=0.2$)} & \multicolumn{4}{c}{$\beta$ ($\alpha=0.3$)} \\
    \cline{3-10}
    Metric & Loss & \textit{0.1} & \textit{0.2} & \textit{0.3} & \textit{0.5} & \textit{0.1} & \textit{0.2} & \textit{0.3} & \textit{0.5} \\
    \midrule
    \midrule
    
    \multirow{3}{*}{Mean MoC} & MSE 
    & 24.0 & 22.0 & 21.4 & 20.6 
    & 29.1 & 26.8 & 25.3 & 24.2 \\
    
    & BCE 
    & \underline{25.6} & \underline{22.8} & \underline{22.4} & \underline{21.9} & 
    \underline{31.7} & \underline{28.9} & \underline{27.2} & \underline{26.1}  \\
    
    & CE &  
    \textbf{26.2} & \textbf{23.3}  & \textbf{23.9} & \textbf{22.6} & 
    \textbf{32.1} & \textbf{29.2} & \textbf{27.5} & \textbf{26.3}  \\
    
    \midrule
    
    \multirow{3}{*}{Top-1 MoC} & MSE 
    & \textbf{51.2} & \textbf{47.3} & \textbf{45.6} & \textbf{45.0}
    & \textbf{54.0} & \textbf{50.4} & \textbf{49.6} & \textbf{47.8} \\
    
    & BCE 
    & \underline{49.4} & \underline{46.0} & \underline{44.6} & \underline{42.6}
    & \underline{52.4} & \underline{48.8} & \underline{47.2} & \underline{44.9} \\
    
    & CE 
    & 46.9 & 42.2 & 41.7 & 40.7
    & 48.8 & 46.5 & 45.3 & 44.0 \\

    \midrule

    \multirow{3}{*}{MFSS} & MSE 
    & \textbf{41.5} & \textbf{44.3} & \textbf{45.6} & \textbf{48.4}
    & \textbf{33.7} & \textbf{36.6} & \textbf{38.0} & \textbf{41.6} \\
    
    & BCE 
    & \underline{35.1} & \underline{37.6} & \underline{38.7} & \underline{39.3}
    & \underline{28.3} & \underline{30.9} & \underline{31.6} & \underline{33.1} \\
    
    & CE 
    & 29.0 & 31.8 & 33.5 & 35.0
    & 24.0 & 26.9 & 28.2 & 30.0 \\

    \bottomrule
    
    \end{tabular}
    }
    \label{tab:stoch_abl_diff_loss_type_breakfast}
\end{table}

\subsubsection{Self conditioning.} 

\begin{figure*}[t!]
    \centering
    \includegraphics[width=1.0\linewidth]{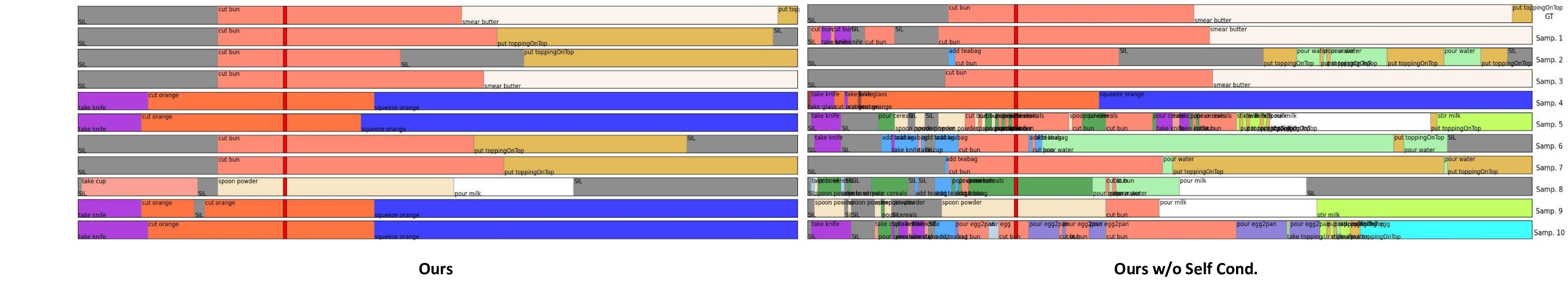}
    \vspace{-0.6cm}
    \caption{Qualitative comparison of the GTD model trained with and without self-conditioning (w/o self cond.). The example is from the Breakfast dataset. Best viewed zoomed in.}
    \label{fig:self_cond}
\end{figure*}

In Tab.~\ref{tab:stoch_abl_self_cond}, we analyze the effect of self-conditioning. Excluding self-conditioning (w/o Self Cond.) decreases the performance. Without self-conditioning, the approach generates many unrealistic predictions. For instance, sample 5 of Fig.~\ref{fig:self_cond} contains many very short action segments and neither the order nor the duration of the actions are plausible.

\begin{table}[t!]
    \caption{
    Ablation on self-conditioning for GTD on Breakfast. 
    Numbers show Mean MoC accuracy.
    }
    \vspace{-0.2cm}
    \centering
    \resizebox{0.65\columnwidth}{!}{%
    \begin{tabular}{l cccc | cccc}
    \toprule
    \multirow{2}{*}{Method} & \multicolumn{4}{c}{$\beta$ ($\alpha=0.2$)} & \multicolumn{4}{c}{$\beta$ ($\alpha=0.3$)} \\
    \cline{2-9}
    & \textit{0.1} & \textit{0.2} & \textit{0.3} & \textit{0.5} & \textit{0.1} & \textit{0.2} & \textit{0.3} & \textit{0.5} \\
    \midrule
    \midrule
    Ours
    & \textbf{24.0} & \textbf{22.0} & \textbf{21.4} & \textbf{20.6}  
    & \textbf{29.1} & \textbf{26.8} & \textbf{25.3} & \textbf{24.2}   \\

    Ours w/o Self Cond. 
    & 23.1 & 21.1 & 20.5 & 19.7 
    & 28.6 & 25.9 & 24.9 & 24.0 \\
    
    \bottomrule
    \end{tabular}
    }
    \label{tab:stoch_abl_self_cond}
\end{table}

\subsubsection{Loss for Observations vs.\ Anticipation.}
In our model, we apply the same loss for the observed frames as well as the future frames. Since the GTD model is only evaluated for anticipation, one might expect that giving a lower weight for the observed frames than for the future frames is beneficial. To this end, we average the loss for the observed frames and future frames and weigh the average over the observed frames by a factor of less than 1. The results are shown in Tab.~\ref{tab:stoch_abl_loss_weight_breakfast}. The best performance for our method is achieved when the weights for both terms are equal ($1.0$) and weighing down the loss applied to the observed frames consistently decreases the performance. 

\begin{table}[t]
    \centering
    \caption{
    Impact of weighting the loss on the observed frames lower than the loss on the future frames for the GTD on Breakfast. In case of $1.0$, observed and future frames are equally weighted. Numbers show Mean MoC accuracy.
    }
    \vspace{-0.2cm}
    \resizebox{0.6\linewidth}{!}{%
    \begin{tabular}{c rrrr | rrrr}
    \toprule
     & \multicolumn{4}{c}{$\beta$ ($\alpha=0.2$)} & \multicolumn{4}{c}{$\beta$ ($\alpha=0.3$)} \\
     \cline{2-9}
    Obs Coef &\textit{0.1} & \textit{0.2} & \textit{0.3} & \textit{0.5} & \textit{0.1} & \textit{0.2} & \textit{0.3} & \textit{0.5} \\
    \midrule
    \midrule
    0.1 & 23.7 & 21.8 & 21.1 & 20.4 & 28.7 & 26.3 & 24.9 & 23.9 \\
    0.5 & 23.8 & 21.8 & 21.2 & \textbf{20.7} & 28.9 & 26.5 & 25.1 & \textbf{24.3} \\
    1.0 & \textbf{24.0} & \textbf{22.0} & \textbf{21.4} & 20.6 & \textbf{29.1} & \textbf{26.8} & \textbf{25.3} & 24.2 \\
    \bottomrule
    \end{tabular}
    }
    \label{tab:stoch_abl_loss_weight_breakfast}
    \end{table}

\subsection{Qualitative Results}
\label{sec:stoch_qual}
\begin{figure*}[h!]
    \centering
    \includegraphics[width=1.0\linewidth]{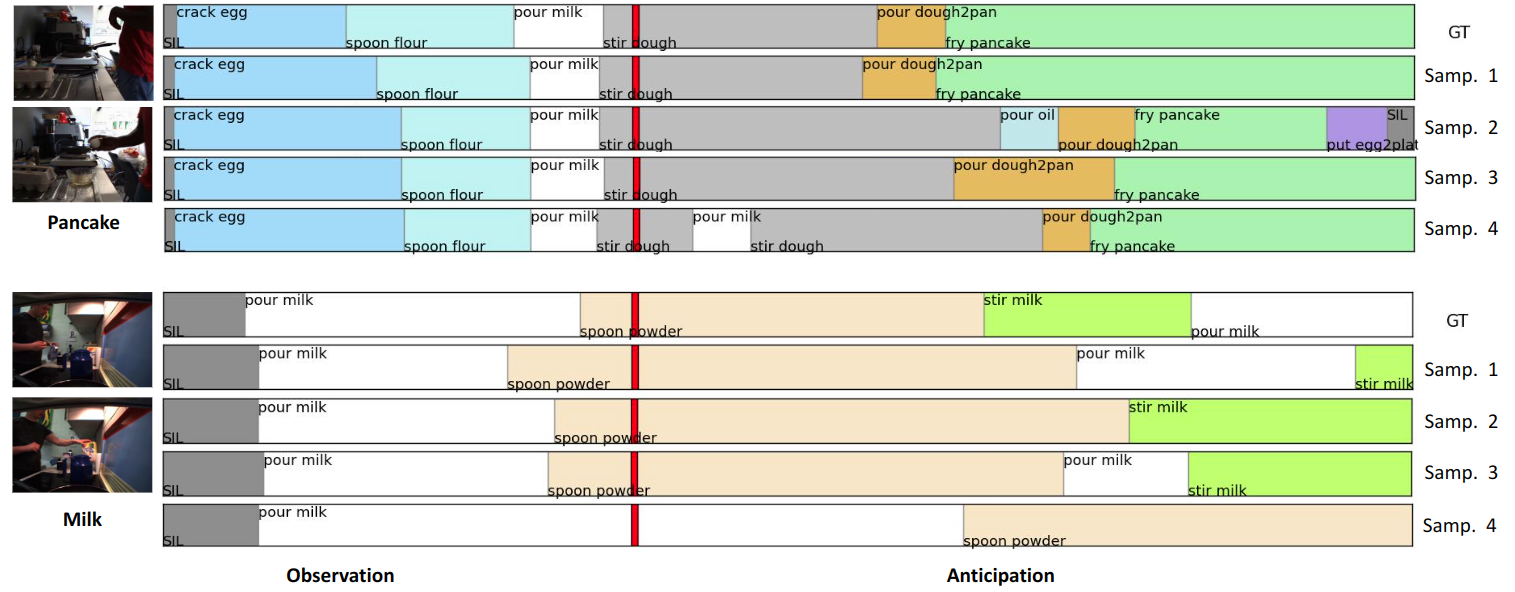}
    \caption{Qualitative results of our proposed GTD for stochastic long-term action anticipation on the Breakfast~\cite{Kuehne12} dataset.}
    \label{fig:prob_bf_1}
    \vspace{-0.5cm}
\end{figure*}

\begin{figure*}[h!]
    \centering
    \includegraphics[width=1.0\linewidth]{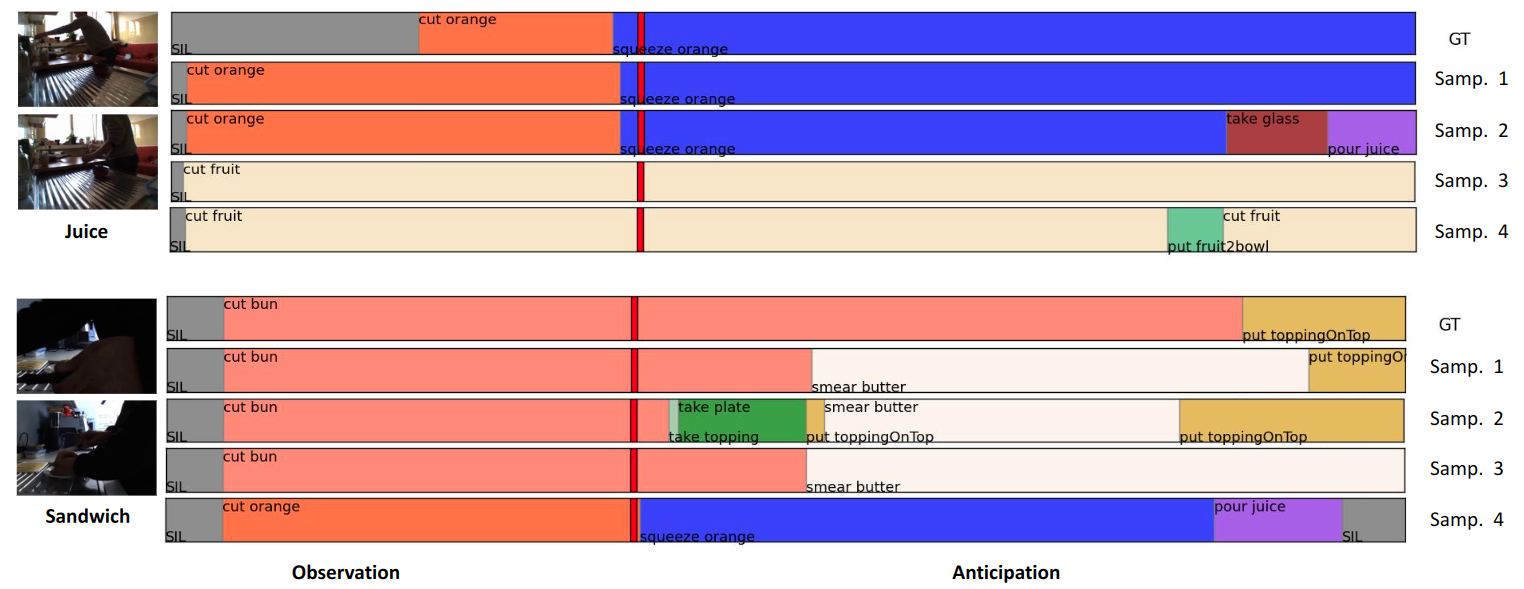}
    \caption{Qualitative results of our proposed GTD for stochastic long-term action anticipation on the Breakfast~\cite{Kuehne12} dataset.}
    \label{fig:prob_bf_2}
    \vspace{-0.5cm}
\end{figure*}

\begin{figure*}[h!]
    \centering
    \includegraphics[width=1.0\linewidth]{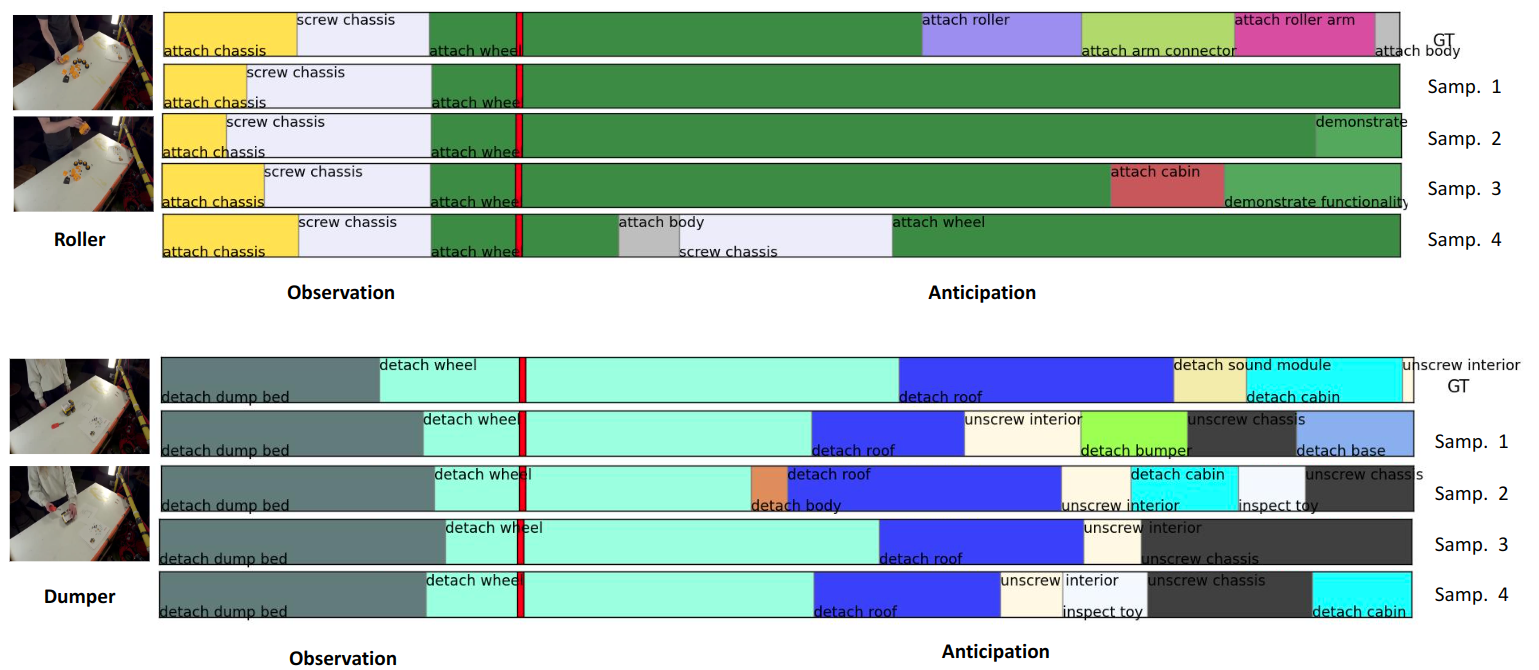}
    \caption{Qualitative results of our proposed GTD for stochastic long-term action anticipation on the Assembly101~\cite{sener2022assembly101} dataset.}
    \label{fig:prob_assembly}
\end{figure*}

\begin{figure*}[h!]
    \centering
    \includegraphics[width=1.0\linewidth]{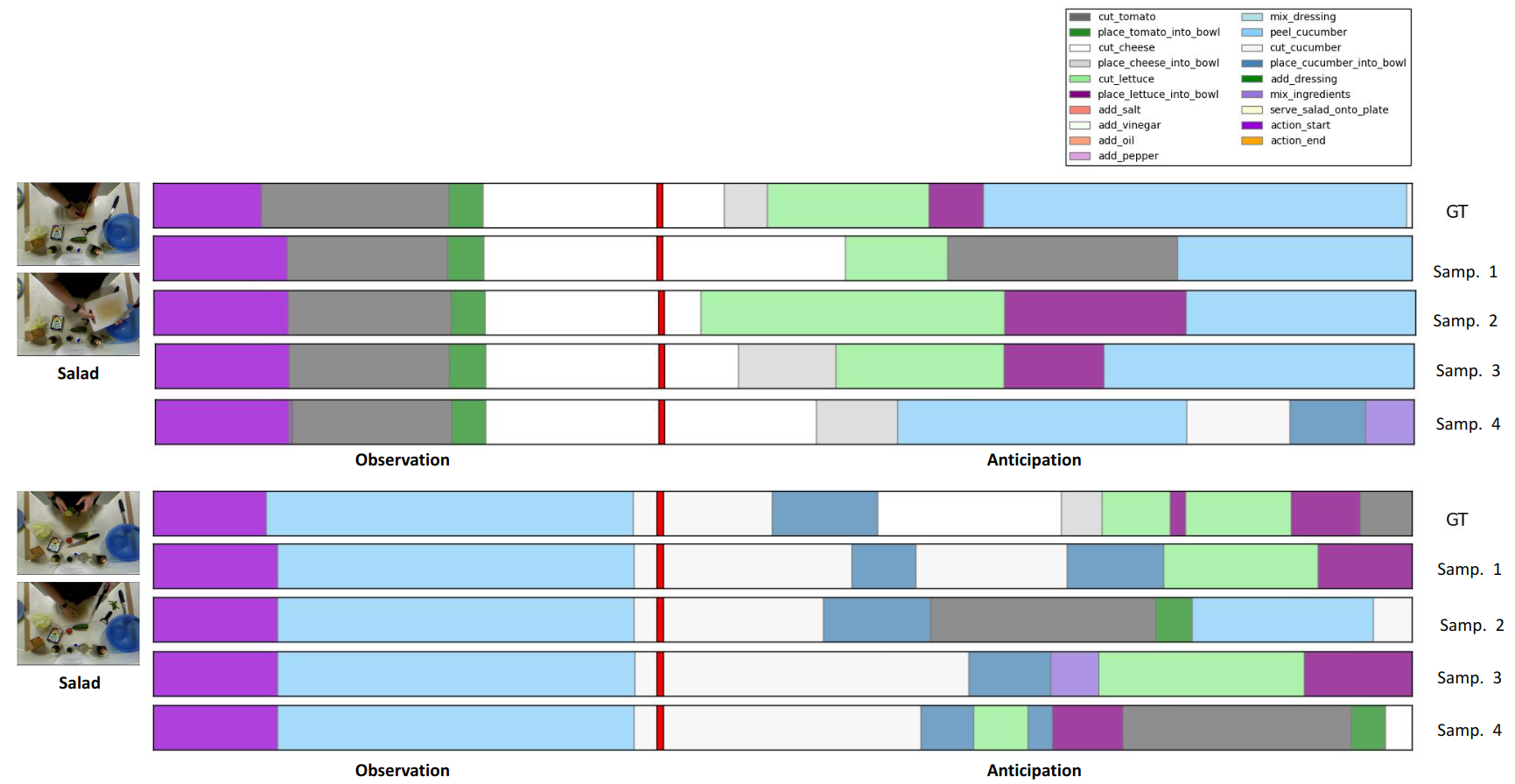}
    \caption{Qualitative results of our proposed GTD for stochastic long-term action anticipation on the 50Salads~\cite{Stein2013CombiningEA} dataset.}
    \label{fig:prob_50s}
\end{figure*}

We present additional qualitative results of GTD in Figs.~\ref{fig:prob_bf_1}
and~\ref{fig:prob_bf_2} on the Breakfast dataset, in Fig.~\ref{fig:prob_assembly} on Assembly101 and in Fig.~\ref{fig:prob_50s} on the 50Salads dataset. Additionally, in Figs.~\ref{fig:all_1}-\ref{fig:all_5}, we present all $M=25$ samples produced by our model on the Breakfast dataset. For each of these examples, we indicate to which observation diversity group they fall, following the grouping from Fig.~\ref{fig:obs_mfss_fut_mfss}.

If the observed frames can be clearly recognized, as in Figs.~\ref{fig:prob_bf_1} and \ref{fig:all_1}, the GTD model predicts sequences from the correct high-level activity with variations in the length and order of individual actions. 
In some cases, however, there are ambiguities in the observed frames due to poor light conditions or occlusions, \eg in Figs.~\ref{fig:prob_bf_2}, \ref{fig:all_3}, \ref{fig:all_4} and \ref{fig:all_5}. In this case, the model generates different interpretations of the observed frames and anticipates actions consistent with the interpretation of the observation. This leads to more diverse predictions for the future, reflecting the uncertainty of the observed frames.  

For example, in Figs.~\ref{fig:prob_bf_2} and \ref{fig:all_4}, the object being cut by the person is barely visible. Consequently, the network generates samples with different interpretations, \textit{i.e.} a bun or an orange as the object being cut.
In Fig.~\ref{fig:all_3}, when the person quickly takes a bowl from a cupboard, it becomes partially occluded by a thermos jug. As a result, the object becomes ambiguous, leading the model to predict actions such as `take bowl' or `take cup' depending on the sample. Similarly, the next object (cereals) remains unclear due to occlusions and motion blur.
Finally, in Fig.~\ref{fig:all_5}, the main action takes place in the corner of the frame due to unfavourable camera positioning. Meanwhile, the foreground features a static cup of coffee, unrelated to the depicted activity of making `juice'. This scenario confuses the model, resulting in predictions ranging from the correct high-level activity to unrelated ones such as `coffee', `tea', or `chocolate milk'. Additionally, the limited visibility of the orange leads to the model interpreting it as an egg for some samples and predicting the sequence from the high-level activity of the `scrambled egg'.

These examples clearly demonstrate that it is essential to model ambiguities of the observation and the future together. The proposed GTD model is the first approach to do so.

\clearpage


{
\begin{figure*}[h!]
    \centering
    \includegraphics[width=1.0\linewidth]{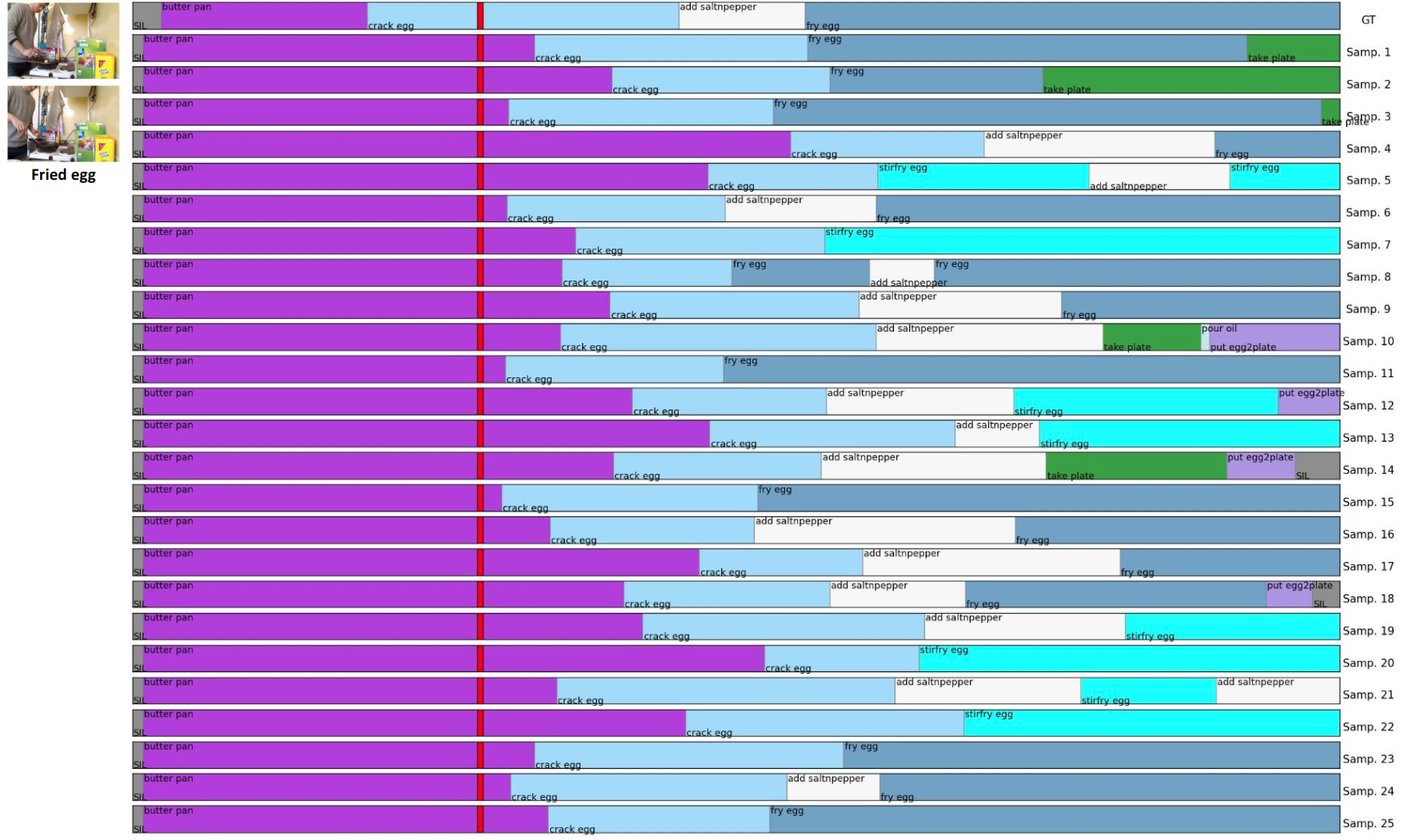}
    \caption{Qualitative results of our proposed GTD for stochastic long-term action anticipation for a video from the first quartile (MFSS for observation) of the Breakfast~\cite{Kuehne12} dataset. Best viewed zoomed in.}
    \label{fig:all_1}
\end{figure*}

\begin{figure*}[h!]
    \centering
    \includegraphics[width=1.0\linewidth]{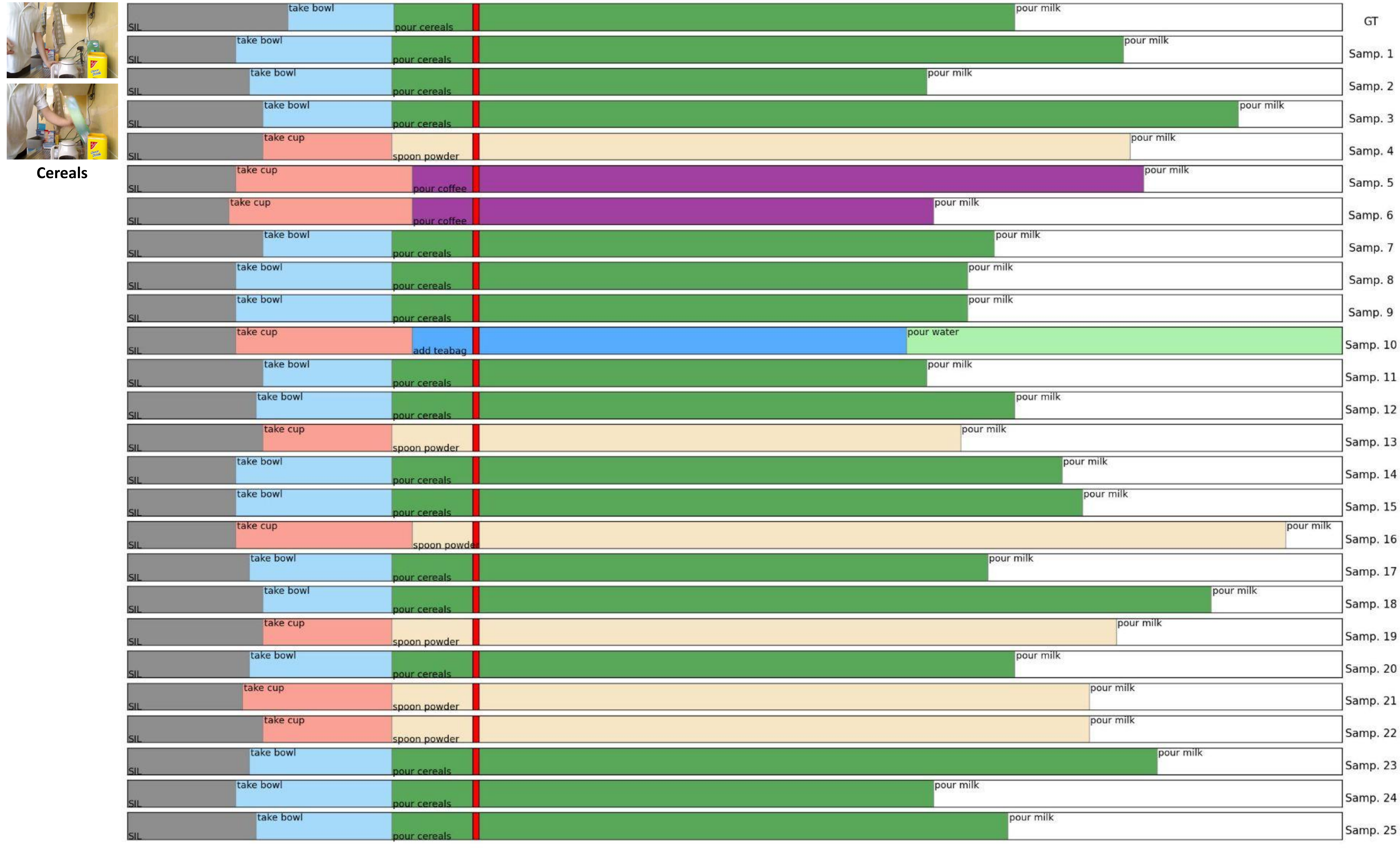}
    \caption{Qualitative results of our proposed GTD for stochastic long-term action anticipation for a video from the second quartile (MFSS for observation) of the Breakfast~\cite{Kuehne12} dataset. Best viewed zoomed in.}
    \label{fig:all_3}
\end{figure*}
}

{
\begin{figure*}[h!]
    \centering
    \includegraphics[width=1.0\linewidth]{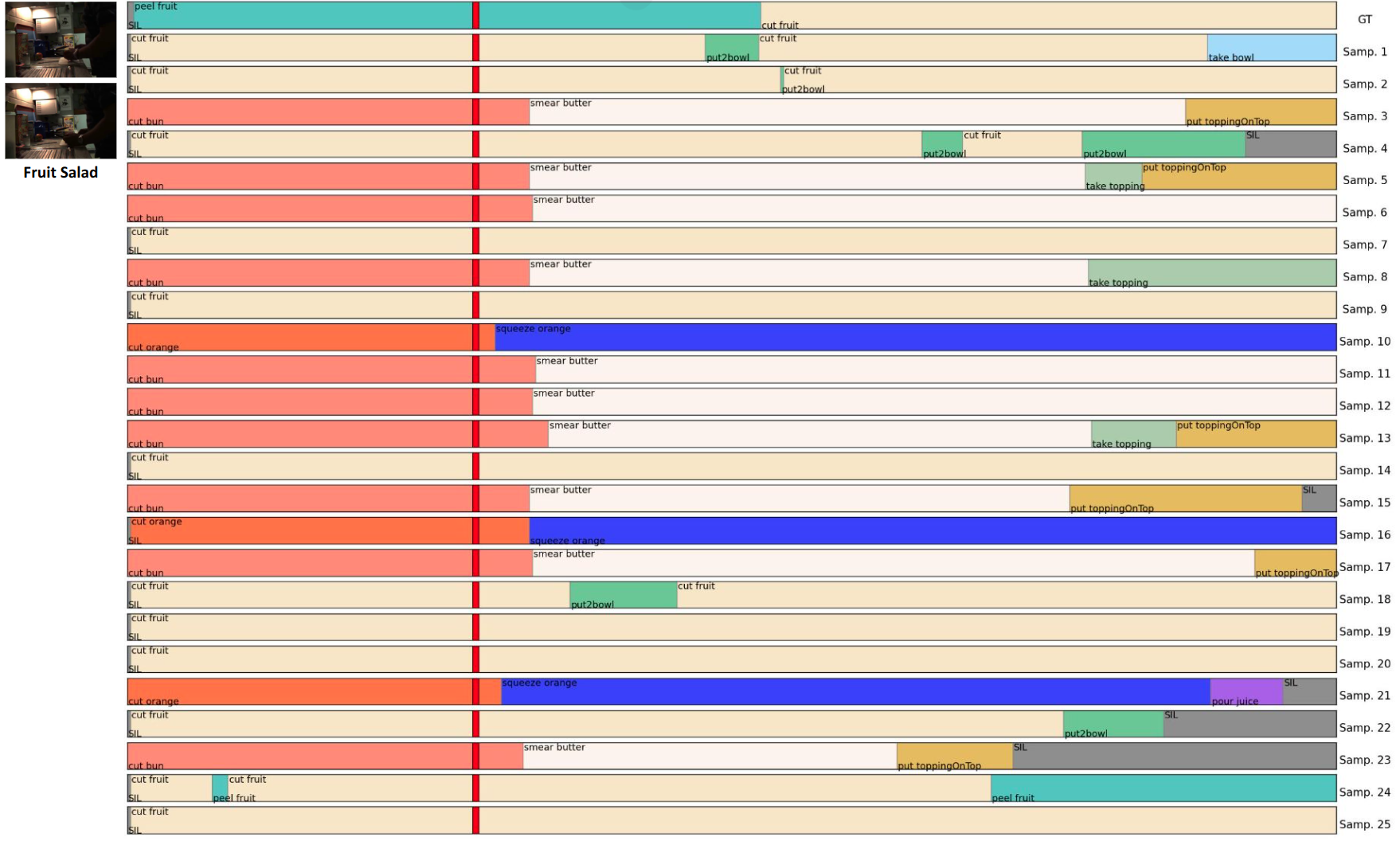}
    \caption{Qualitative results of our proposed GTD for stochastic long-term action anticipation for a video from the third quartile (MFSS for observation) of the Breakfast~\cite{Kuehne12} dataset. Best viewed zoomed in.}
    \label{fig:all_4}
\end{figure*}

}

{
\begin{figure*}[h!]
    \centering
    \includegraphics[width=1.0\linewidth]{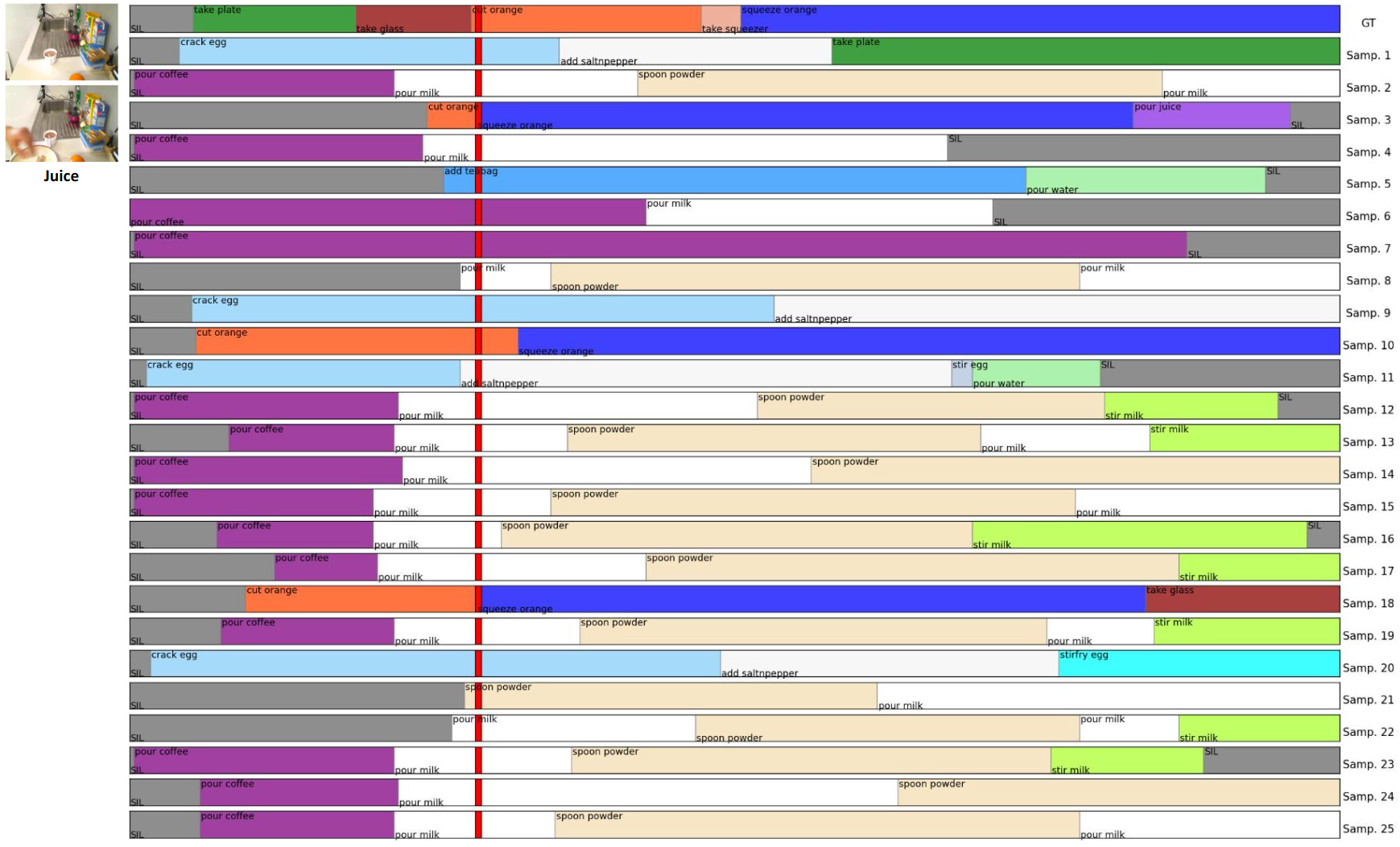}
    \caption{Qualitative results of our proposed GTD for stochastic long-term action anticipation for a video from the fourth quartile (MFSS for observation) of the Breakfast~\cite{Kuehne12} dataset. Best viewed zoomed in.}
    \label{fig:all_5}
\end{figure*}

}
\clearpage

\section{Deterministic Long-Term Anticipation}
\label{sec:determ}

\subsection{Comparison to State of The Art}
We present the results of our proposed GTAN model for deterministic long-term action anticipation on the 50Salads dataset in Tab.~\ref{tab:results_main_determ_50s}. On this dataset, there is no clear leading approach. While \cite{Ke_2019_CVPR} performs best in most cases for $\alpha=0.3$, our approach performs best for most cases for $\alpha=0.2$. Overall, our model performs on par with the state of the art on this dataset.

\begin{table}[h!]
\vspace{-0.2cm}
\centering
\caption{
Comparison with state-of-the-art methods for deterministic anticipation on the 50Salads dataset. ** Indicates results from the FUTR github repository.
}
\vspace{-0.3cm}
\resizebox{0.8\linewidth}{!}{%
\setlength{\tabcolsep}{3.5pt}
\begin{tabular}{ll rrrr | rrrr}
\toprule
\multirow{2}{*}{Dataset} & \multirow{2}{*}{Method}  & \multicolumn{4}{c}{$\beta$ ($\alpha=0.2$)} & \multicolumn{4}{c}{$\beta$ ($\alpha=0.3$)} \\
\cline{3-10}
& & \textit{0.1} & \textit{0.2} & \textit{0.3} & \textit{0.5} & \textit{0.1} & \textit{0.2} & \textit{0.3} & \textit{0.5} \\

\midrule
\midrule
\multirow{8}{*}{50Salads} & RNN~\cite{Farha_2018_CVPR}
& 30.1 & 25.4 & 18.7 & 13.5 
& 30.8 & 17.2 & 14.8 & 9.8 \\

& CNN~\cite{Farha_2018_CVPR} 
& 21.2 & 19.0 & 15.9 & 9.9
& 29.1 & 20.1 & 17.5 & 10.9 \\

& UAAA (mode)~\cite{farha2019uaaa}
& 24.9 & 22.4 & 19.9 & 12.8
& 29.1 & 20.5 & 15.3 & 12.3 \\

& Time Cond.~\cite{Ke_2019_CVPR}
& 32.5 & 27.6 & 21.3 & 16.0
& \textbf{35.1} & \textbf{27.1} & \textbf{22.1} & \underline{15.6} \\

& TempAgg~\cite{sener2020temporal} 
& 25.5 & 19.9 & 18.2 & 15.1
& 30.6 & 22.5 & \underline{19.1} & 11.2 \\

& Cycle Cons.~\cite{farha2020gcpr} 
& 34.8 & \textbf{28.4 }& 21.8 & 15.3
& \underline{34.4} & 23.7 & 19.0 & \textbf{15.9} \\

& FUTR**~\cite{gong2022future} 
& \textbf{37.0} & \underline{27.8} & \underline{22.5} & 16.8
& {33.3} & 23.2 & \textbf{22.1} & 15.5 \\

& \textbf{Ours}                      
& \underline{36.7} & 27.7 & \textbf{23.8} & \textbf{17.4}
& 32.2 & \underline{24.9} &  17.4 & 14.9 \\
\bottomrule
\end{tabular}
}
\label{tab:results_main_determ_50s}
\vspace{-0.3cm}
\end{table}

\subsection{Ablations}

\label{sec:abl_determ}
In this section, we provide a set of ablation experiments for our proposed model in the deterministic setting. 

\subsubsection{Number of Stages and Layers.} We analyze the impact of the number of stages and layers in the GTAN model. The results reported in Tab.~\ref{tab:determ_abl_num_stages_layers_breakfast} show that the best performance is achieved by the model with $S=5$ stages and $L=9$ layers. Additionally, we analyze if gated convolutions are required at all stages of the network. We did so by using them only in the first $S'$ stages while using only feature convolutions for the remaining stages. 
The respective results with different $S'$ are shown in Tab.~\ref{tab:determ_abl_gated_stages_breakfast}. 
The results show that using gated convolutions for all stages performs best.  
As mentioned before, the differentiation between the observed and anticipated parts is preserved even in the later stages of the proposed GTAN model, which hints at the usefulness of such separation. Therefore, we speculate that the deterioration of the results for $S'<S$ was caused by the inability of the network's latest stages to distinguish between the observed and future frames caused by the absence of the gating mechanism.

\begin{table}[h!]
\centering
\caption{
Ablation on the number of layers and stages for GTAN on Breakfast.
}
\vspace{-0.2cm}
\resizebox{0.65\columnwidth}{!}{%
\begin{tabular}{ll rrrr | rrrr}
\toprule
Num. & Num. & \multicolumn{4}{c}{$\beta$ ($\alpha=0.2$)} & \multicolumn{4}{c}{$\beta$ ($\alpha=0.3$)} \\
 \cline{3-10}
Layers & Stages &\textit{0.1} & \textit{0.2} & \textit{0.3} & \textit{0.5} & \textit{0.1} & \textit{0.2} & \textit{0.3} & \textit{0.5} \\
\midrule
\midrule
9 & 4 & 27.5 & 24.8 & 24.1 & 24.0 & 32.1 & 29.6 & 28.3 & 28.2 \\
9 & 5 & 28.8 & \textbf{26.3} & \textbf{25.8} &\textbf{ 26.0} & \textbf{35.5} & \textbf{32.9} & \textbf{30.5} & \textbf{29.6} \\
9 & 6 & \textbf{29.3} & 25.9 & 25.0 & 25.1 & 33.7 & 31.0 & 29.5 & 29.0 \\
\midrule
\midrule
8 & 5 & 27.7 & 24.6 & 24.1 & 23.9 & 33.1 & 30.1 & 28.5 & 28.3 \\
9 & 5 & \textbf{28.8} & \textbf{26.3} & \textbf{25.8} & \textbf{26.0} & \textbf{35.5} & \textbf{32.9} & \textbf{30.5} & \textbf{29.6} \\
10 & 5 & 27.8 & 25.3 & 24.4 & 24.1 & 32.0 & 29.3 & 27.5 & 27.2 \\
\bottomrule
\end{tabular}
}
\label{tab:determ_abl_num_stages_layers_breakfast}
\end{table}

\begin{table}[h!]
\centering
\vspace{-0.1cm}
\caption{
Ablation on the number of stages with gated convolutions on Breakfast.
}
\vspace{-0.2cm}
\resizebox{0.5\columnwidth}{!}{%
\begin{tabular}{l cccc | cccc}
\toprule
 Stages & \multicolumn{4}{c}{$\beta$ ($\alpha=0.2$)} & \multicolumn{4}{c}{$\beta$ ($\alpha=0.3$)} \\
\cline{2-9}
 w. gated &\textit{0.1} & \textit{0.2} & \textit{0.3} & \textit{0.5} & \textit{0.1} & \textit{0.2} & \textit{0.3} & \textit{0.5} \\
\midrule
\midrule
1 / 5& 28.7 & 25.6 & 24.6 & 24.1 & 33.7 & 31.3 & 29.2 & 27.9 \\
3 / 5& \textbf{28.8} & \textbf{26.4} & 25.7 & 25.2 & 34.3 & 32.1 & 29.5 & 28.6 \\
5 / 5 & \textbf{28.8} & 26.3 & \textbf{25.8} & \textbf{26.0} & \textbf{35.5} & \textbf{32.9} & \textbf{30.5} & \textbf{29.7} \\
\bottomrule
\end{tabular}
}
\vspace{-0.3cm}
\label{tab:determ_abl_gated_stages_breakfast}
\end{table}

\subsubsection{Gated Temporal Convolution.}
We analyze the effect of gated convolutions on our deterministic GTAN model. To this end, we performed the same series of experiments as for GTD, shown in Tab.~\ref{tab:determ_abl_gated_breakfast}. 
Removing the gate convolution branch and leaving only the feature branch with standard temporal convolutions (Ours w/o GC) led to considerably lower performance compared to the proposed approach. Also removing the dilation from the gated convolution (Ours w/o Dil. GC) resulted in the performance drop. This demonstrates the necessity of using the gating mechanism. 
We also tested three alternative gating formulations - Aslan~\etal~\cite{aslan2022gtcn}, SE~\cite{Hu2017SqueezeandExcitationN} and Partial Convolution~\cite{Liu2018ImageIF}. The performance of these methods is inferior to the suggested approach. The results are consistent with the results in Tab.~3 of the main paper.

\begin{table}[h!]

\centering
\caption{
Ablation on the gated convolution for GTAN on Breakfast.}
\vspace{-0.2cm}
\resizebox{0.65\columnwidth}{!}{%
\begin{tabular}{lrrrrrrrr}
\toprule
 & \multicolumn{4}{c}{$\beta$ ($\alpha=0.2$)} & \multicolumn{4}{c}{$\beta$ ($\alpha=0.3$)} \\
\cline{2-9}
 &\textit{0.1} & \textit{0.2} & \textit{0.3} & \textit{0.5} & \textit{0.1} & \textit{0.2} & \textit{0.3} & \textit{0.5} \\
\midrule
Ours & 28.8 & \textbf{26.3 }& \textbf{25.8 }& \textbf{26.0} & \textbf{35.5} & \textbf{32.9} & \textbf{30.5} & \textbf{29.6} \\
\midrule
\midrule
Ours w/o GC & 28.1 & 24.6 & 24.0 & 23.9 & 32.7 & 30.3 & 28.3 & 27.3 \\
Ours w/o Dil. GC & \textbf{ 28.9} & 25.8 & 24.9 & 25.4 & 33.3 & 30.5 & 29.2 & 28.5 \\
Aslan~\etal~\cite{aslan2022gtcn} & 21.9 & 20.7 & 19.7 & 19.0 & 25.8 & 23.1 & 22.4 & 20.0 \\

\midrule
Part. Conv~\cite{Liu2018ImageIF} & 28.8 & 25.3 & 24.6 & 24.1 & 32.9 & 30.5 & 28.2 & 27.6 \\
SE~\cite{Hu2017SqueezeandExcitationN} & 27.2 & 24.7 & 24.6 & 23.3 & 31.7 & 28.9 & 27.4 & 27.9 \\
\bottomrule
\end{tabular}
}

\label{tab:determ_abl_gated_breakfast}
\end{table}

\textbf{Loss Type.} As for the GTD model, we have analyzed the effect of the training loss on our proposed GTAN model in the deterministic setting. Instead of using the cross-entropy (CE) loss, we have trained our network with the mean-square-error (MSE) loss. The results reported in Tab.~\ref{tab:determ_abl_loss_type_breakfast} show that using the MSE loss in the deterministic setting leads to a considerable decrease in the performance.

\begin{table}[h!]
\vspace{-0.3cm}
\centering
\caption{
Impact of using the MSE loss instead of the cross entropy (CE) loss in the deterministic setting. 
}
\vspace{-0.2cm}
\resizebox{0.55\columnwidth}{!}{%
\begin{tabular}{l cccc | cccc}
\toprule
\multirow{2}{*}{Loss} & \multicolumn{4}{c}{$\beta$ ($\alpha=0.2$)} & \multicolumn{4}{c}{$\beta$ ($\alpha=0.3$)} \\
 \cline{2-9}
 &\textit{0.1} & \textit{0.2} & \textit{0.3} & \textit{0.5} & \textit{0.1} & \textit{0.2} & \textit{0.3} & \textit{0.5} \\
\midrule
\midrule
MSE & 23.8 & 21.8 & 21.1 & 19.3 & 27.4 & 25.5 & 24.2 & 22.2 \\
CE & \textbf{28.8} & \textbf{26.3} & \textbf{25.8} & \textbf{26.0} & \textbf{35.5} & \textbf{32.9}  & \textbf{30.5} & \textbf{29.6} \\
\bottomrule
\end{tabular}
}
\vspace{-0.05in}
\label{tab:determ_abl_loss_type_breakfast}
\end{table}

\subsection{Qualitative Results}
\begin{figure*}[h!]
    \centering
    \includegraphics[width=1.0\linewidth]{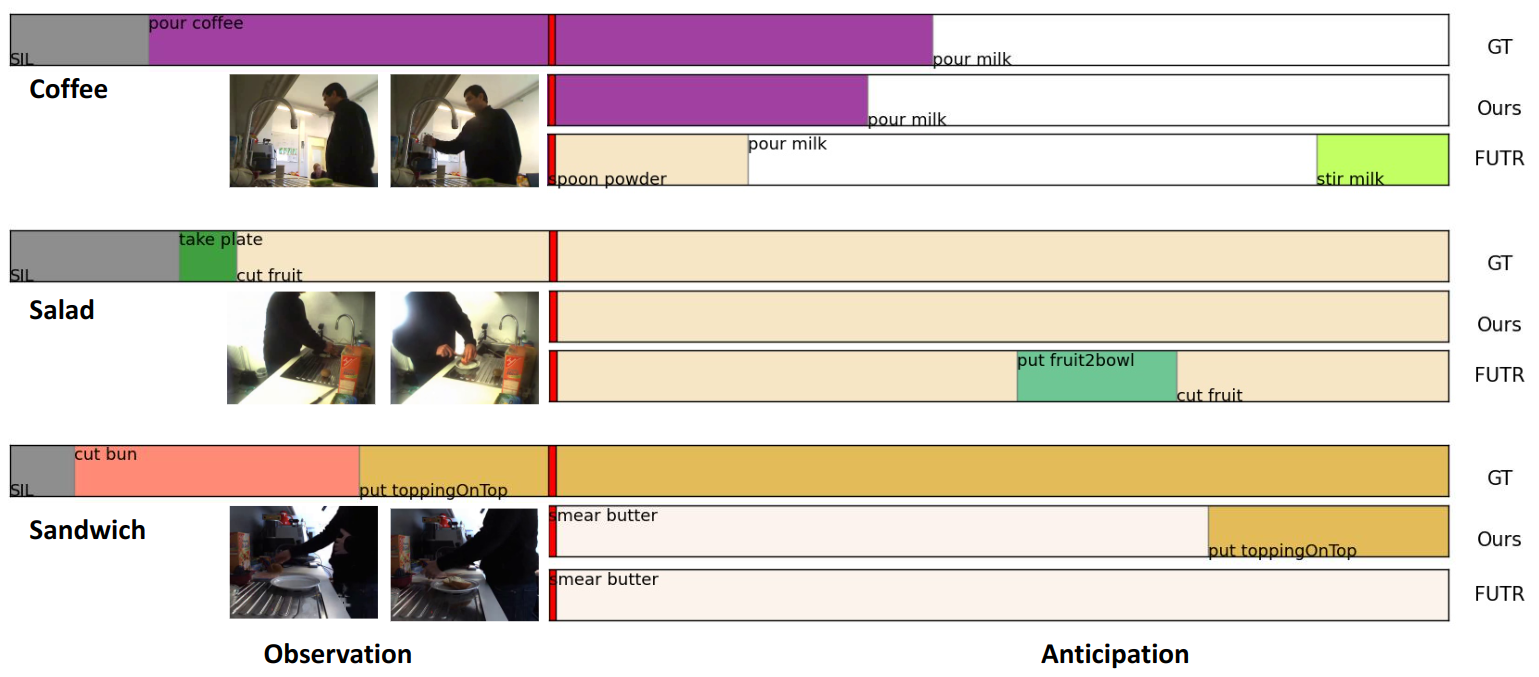}
    \caption{Qualitative comparisons of our approach with FUTR~\cite{gong2022future} in the deterministic setting on the Breakfast~\cite{Kuehne12} dataset.}
    \label{fig:determ_bf_1}
\end{figure*}

\begin{figure*}[h!]
    \centering
    \includegraphics[width=1.0\linewidth]{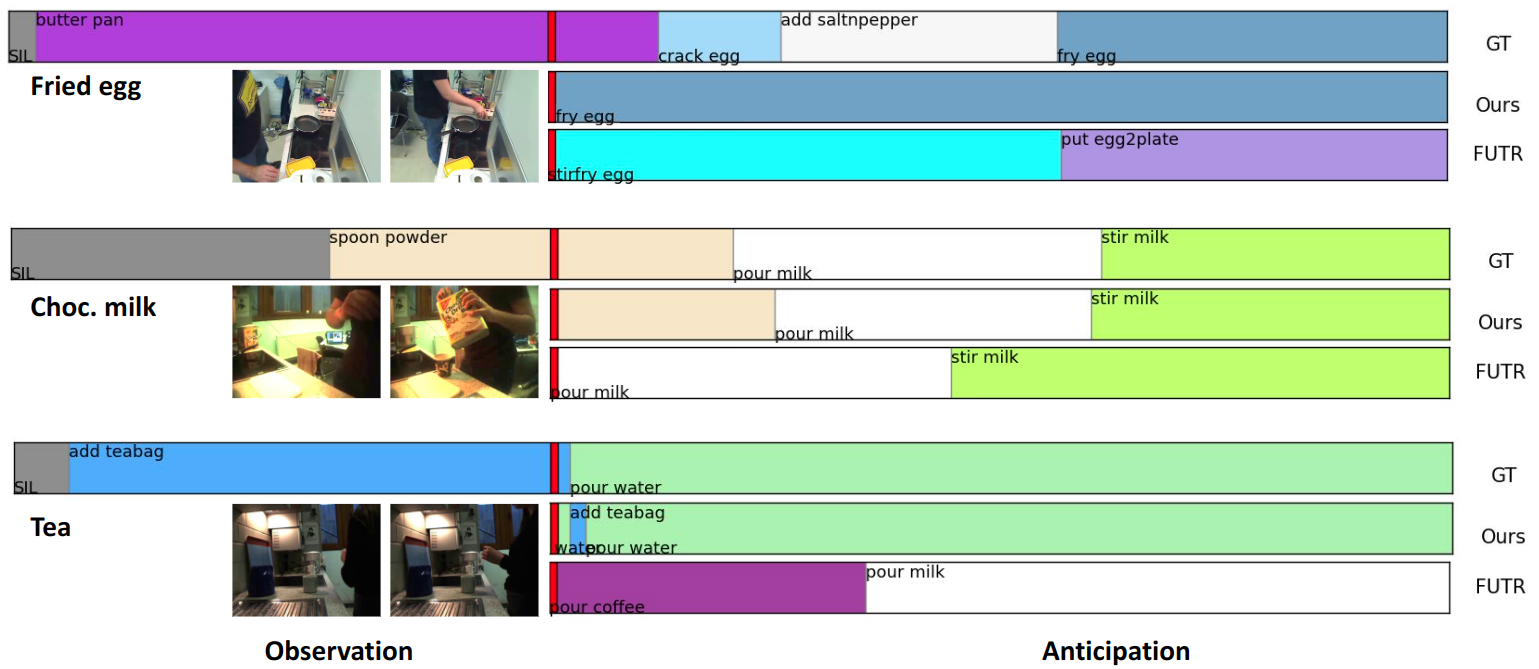}
    \caption{Qualitative comparisons of our approach with FUTR~\cite{gong2022future} in the deterministic setting on the Breakfast~\cite{Kuehne12} dataset.}
    \label{fig:determ_bf_2}
\end{figure*}

\begin{figure*}[h!]
    \centering
    \includegraphics[width=1.0\linewidth]{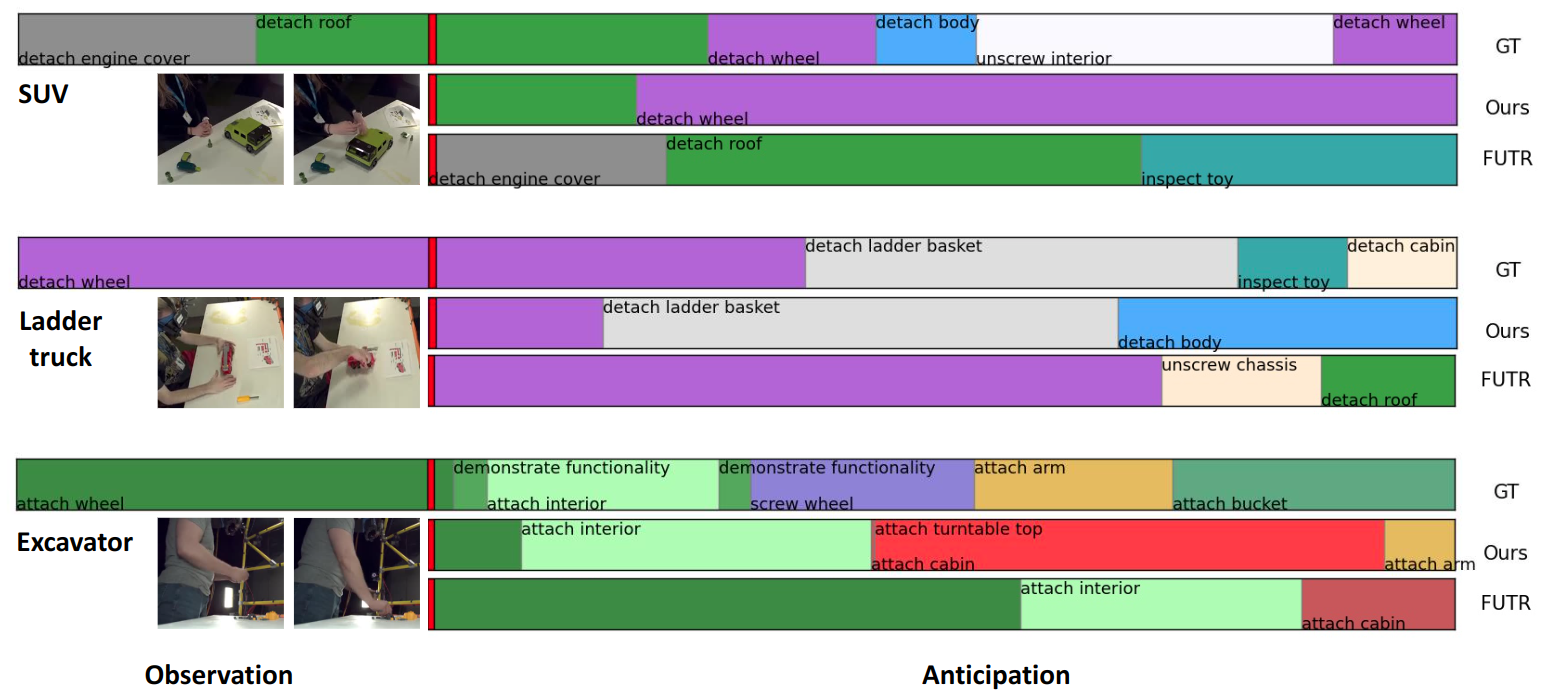}
    \caption{Qualitative comparisons of our approach with FUTR~\cite{gong2022future} in the deterministic setting on the Assembly101~\cite{sener2022assembly101} dataset.}
    \label{fig:determ_assembly_1}
\end{figure*}

\begin{figure*}[h!]
    \centering
    \includegraphics[width=1.0\linewidth]{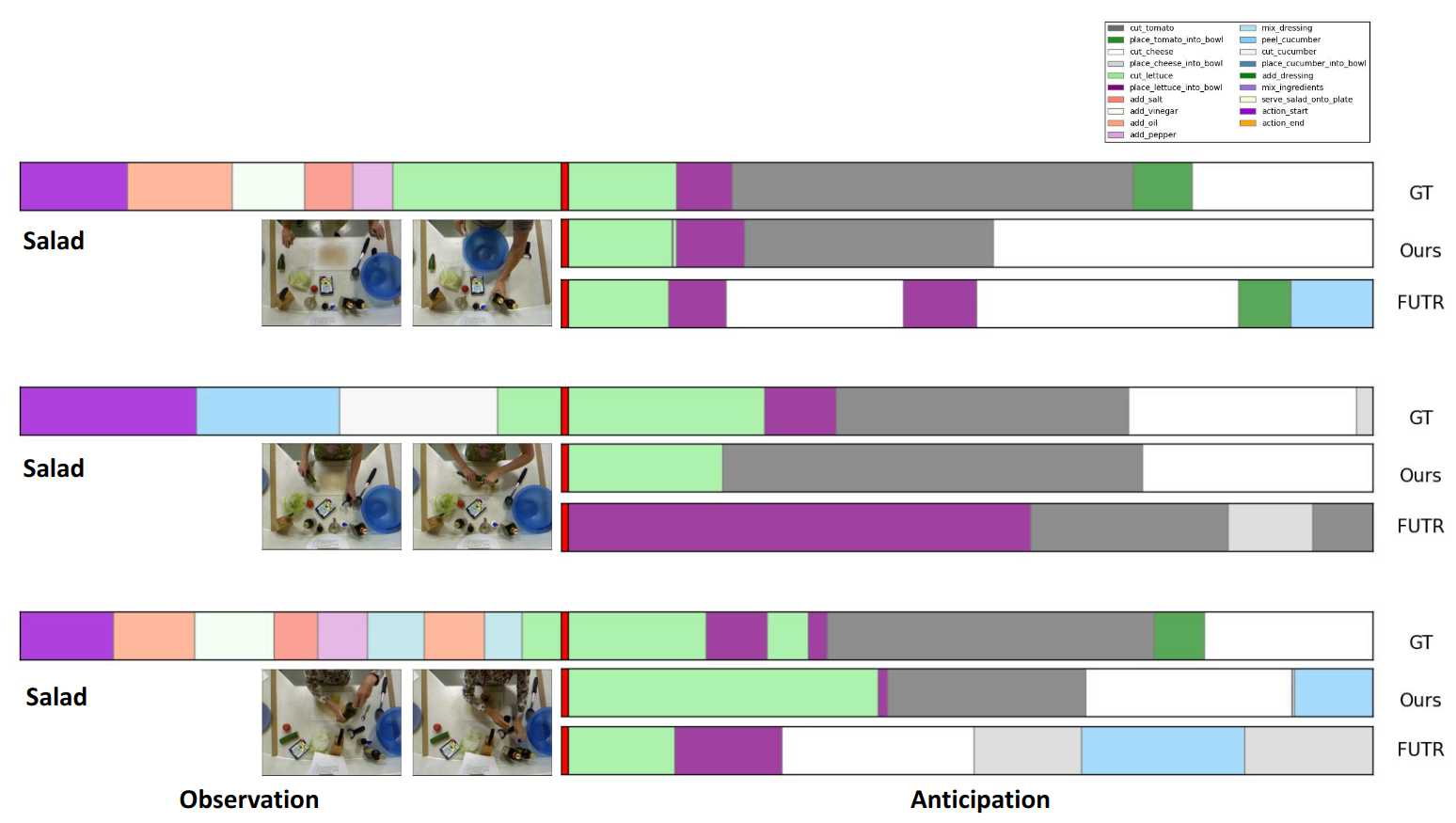}
    \caption{Qualitative comparisons of our approach with FUTR~\cite{gong2022future} in the deterministic setting on the 50Salads~\cite{Stein2013CombiningEA} dataset.}
    \label{fig:determ_50s}
\end{figure*}

We present additional qualitative results of our approach in the deterministic setting in Figs.~\ref{fig:determ_bf_1} and~\ref{fig:determ_bf_2} on the Breakfast dataset, in Fig.~\ref{fig:determ_assembly_1} on Assembly101 and in Fig.~\ref{fig:determ_50s} on 50Salads dataset. 
In the first example of Fig.~\ref{fig:determ_bf_1} and the first and third examples of Fig.~\ref{fig:determ_bf_2}, FUTR fails to recognize the correct high-level actions, which leads to the overall wrong predictions of the upcoming actions. In other cases, the high-level action is recognized correctly, however, the duration or presence of some actions is anticipated inaccurately.

\clearpage


%
%
\clearpage
\bibliographystyle{splncs04}
\bibliography{main}
\end{document}